\documentclass[10pt,twocolumn,letterpaper]{article}

\usepackage[pagenumbers]{iccv} 
\usepackage[accsupp]{axessibility}
\usepackage{url}
\usepackage{mathtools}
\usepackage{pifont}
\usepackage{multirow}
\usepackage{subcaption}

\usepackage{graphicx} 
\usepackage{booktabs}

\definecolor{iccvblue}{rgb}{0.21,0.49,0.74}
\usepackage[pagebackref,breaklinks,colorlinks,allcolors=iccvblue]{hyperref}

\def\paperID{9119}
\def\confName{ICCV}
\def\confYear{2025}

\usepackage{etoolbox}
\newcommand{\insertfig}{\vspace{0.3cm}\includegraphics[width=\textwidth]{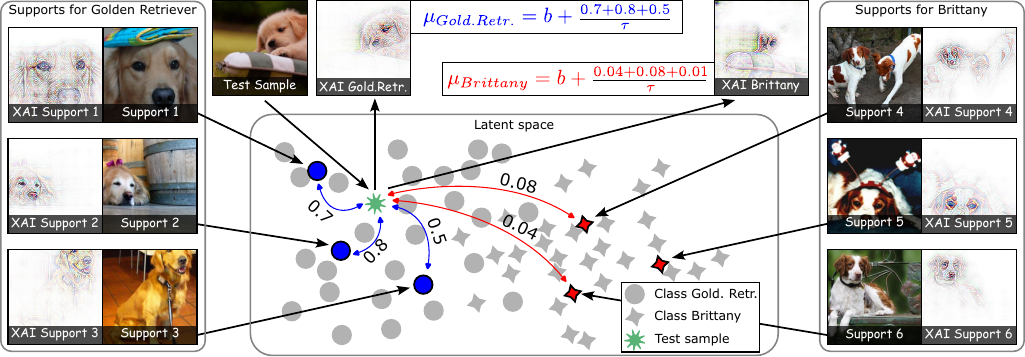}
    \captionof{figure}{
    \ourmod\ trains a neural network to extract class-representative support feature vectors (red and blue) from training images.
    It computes class logits $\mu$ as the sum of temperature-normalized similarity scores between a class's support feature vectors and a test sample's feature vector (green), as shown for the predicted $Golden~Retriever$ class and non-predicted $Brittany$ class. Its B-cos backbone permits the computation of faithful local and global explanations. Stars and circles denote samples of different classes.}
    \label{fig:fig1}
    \vspace{0.5cm}}
\makeatletter
\apptocmd{\@maketitle}{\centering\insertfig}{}{}
\makeatother

\title{SIC: Similarity-Based Interpretable Image Classification with Neural Networks}

\author{Tom Nuno Wolf \and Emre Kavak \and Fabian Bongratz \and Christian Wachinger
\and
Lab for Artificial Intelligence in Medical Imaging, Technical University of Munich, Germany
\and
Munich Center for Machine Learning (MCML), Germany\\
{\tt\small {tn.wolf}@tum.de}
}

\newcommand{\cmark}{\ding{51}}
\newcommand{\xmark}{\ding{55}}

\newcommand{\ourmod}{SIC}

\begin{document}

\maketitle

\begin{abstract}
The deployment of deep learning models in critical domains necessitates a balance between high accuracy and interpretability.
We introduce \ourmod, an inherently interpretable neural network that provides local and global explanations of its decision-making process.
Leveraging the concept of case-based reasoning, \ourmod\ extracts class-representative support vectors from training images, ensuring they capture relevant features while suppressing irrelevant ones.
Classification decisions are made by calculating and aggregating similarity scores between these support vectors and the input's latent feature vector. 
We employ B-Cos transformations, which align model weights with inputs, to yield coherent pixel-level explanations in addition to global explanations of case-based reasoning.
We evaluate \ourmod\ on three tasks: fine-grained classification on Stanford Dogs and FunnyBirds, multi-label classification on Pascal VOC, and pathology detection on the RSNA dataset.
Results indicate that \ourmod\ not only achieves competitive accuracy compared to state-of-the-art black-box and inherently interpretable models but also offers insightful explanations verified through practical evaluation on the FunnyBirds benchmark.
Our theoretical analysis proves that these explanations fulfill established axioms for explanations.
Our findings underscore \ourmod's potential for applications where understanding model decisions is as critical as the decisions themselves.
\end{abstract}


\section{Introduction}
Deep learning is increasingly becoming part of everyday life, e.g., targeted advertisements, gaming, and high-stakes areas like banking and medicine.
In the latter, identifying errors in the model’s decision-making process is crucial to detect failure cases, which can be addressed by adding transparency to model predictions using explainable AI (XAI).
Explanations are typically produced post-hoc for black-box models using approximation methods, which estimate the contribution of each input feature to the prediction~\citep{bach2015pixel,ribeiro2016should,selvaraju2017grad,Lundberg_2017,axioms,petsiuk2018rise}.
While these methods can accurately compute feature contributions for low-dimensional inputs~\cite{Lundberg_2017}, they introduce  approximation errors for high-dimensional inputs such as images~\citep{adebayo2018sanity}.

As an alternative, \emph{inherently interpretable models} provide explanations by design. 
For instance, logistic regression can be considered an inherently interpretable model since evaluating its coefficients directly explains the effect of each input feature towards the prediction.
Translating these properties to complex functions implemented by neural networks was further fueled by the argument that if a technique is able to summarize the decision-making of a black-box without any error, the black-box could be replaced by an inherently interpretable model in the first place~\citep{Rudin_2019}.
As a result, a growing research community is dedicated to developing inherently interpretable models, e.g., for decision-critical tasks like medicine~\citep{barnett2021interpretable,kim2021xprotonet,Wolf_2023}.

The explanations provided by these models can be categorized into local and global explanations, and evaluated in terms of axioms to verify if they are faithful~\citep{axioms}~\footnote{Throughout this manuscript, we define faithful explanations as explanations that satisfy all of the axioms proposed by~\citet{axioms}: Completeness, Sensitivity, Implementation Invariance, Dummy, Linearity, and Symmetry-Preserving.}.
Local explanations focus on individual predictions, revealing how the model arrived at a specific outcome for a particular instance. In contrast, global explanations provide insight into the model's overall behavior.  
In logistic regression, global explanations are derived from the model's coefficients, which describe how each feature influences the prediction across all instances. To produce local explanations, a sample’s specific feature values are multiplied by their corresponding coefficients, illustrating how each feature contributed to the particular prediction.

\paragraph{Related Work}
As outlined in Tab.~\ref{tab:properties}, \emph{locally} interpretable deep learning models include the work by~\citet{nwhead_interpret}, who trained a non-parametric classification head with metric learning to classify an image by summing the softmax-normalized similarity scores between training image latent feature vectors and the test image latent feature vector.
They proposed to provide local explanations by showing the most influential images alongside their corresponding scores for a given test sample, similar to the k-nearest-neighbor algorithm.
While intuitive, the approach lacks contribution maps in pixel space to identify important input features.
To this end, BagNet~\citep{brendel2019approximating} classify local patches of an image individually and sum the probabilities of all patches for the final classification.
Providing the class-probability maps of the individual patches' predictions as explanations shows which local structures showed evidence of a respective class.
However, they lack faithful fine-grained attributions computed on a pixel-level, which are provided by B-cos networks \citep{mboehle,mboehle_trans}.
In these models, each forward pass is summarized as a linear transformation, and its resulting weight matrix then serves to compute local explanations in image space. 
However, B-cos models lack global interpretability because the weight matrix is dynamically computed for each input.

\begin{table}[h]
    \centering
    \begin{tabular}{lccc}
        \toprule
        \textbf{Model} & \textbf{Local} & \textbf{Global} & \textbf{Faithful} \\ 
        \midrule
        NW-Head~\cite{nwhead_interpret} & \cmark & \cmark & \xmark \\ 
        ProtoPNet~\cite{chen2019looks} & \cmark & \cmark & \xmark \\ 
        BagNet~\cite{brendel2019approximating} & \cmark & \xmark & \cmark \\ 
        B-Cos~\cite{mboehle} & \cmark & \xmark & (\cmark) \\ 
        \ourmod\ (ours) & \cmark & \cmark & \cmark \\ 
        \bottomrule
    \end{tabular}
    \caption{Comparison of the ability of inherently interpretable models to provide local, global, and faithful explanations.}
    \label{tab:properties}
\end{table}

Models that provide global interpretability include concept bottleneck models~\citep{CBM}, which train an encoder to predict concepts present in an image and a classifier to predict the target class from these concepts, which serve as local explanations during a prediction.
However, they require access to concepts, i.e., meta information on each concept present in an image, e.g., the color of the tail, legs, or head of a bird.
While the concepts themselves provide global and local explanations and are manually pre-defined, concept bottleneck models lack explanations in the image domain. 
The most prominent inherently interpretable model is ProtoPNet~\citep{chen2019looks} and its extensions~\citep{kim2021xprotonet,Donnelly_2022_CVPR}, which provide local and global explanations. 
Its key idea is to learn class-specific prototypical parts across the training set to facilitate reasoning as ``this part of an unseen test image looks like that part from my training images.'' 
To provide image space explanations, it upsamples distance maps from the latent space into the input domain.
However,~\citet{hoffmann2021looks} found that the explanations provided by ProtoPNet can be misleading and called for more rigorous evaluations.
~\citet{sacha2024interpretability} proposed a benchmark to evaluate spatial misalignment of explanations given by ProtoPNet, which was also found by the theoretical evaluation of~\citet{wolf2024keep}. They showed that spatial relationship between the input space and latent space can not be guaranteed for ProtoPNets with deep convolutional neural networks encoders, generalizing to other prototypical networks like Pip-Net~\citep{nauta2023pip}, ProtoTrees~\citep{nauta2021prototree}, XProtoNet~\citep{kim2021xprotonet}, or the Deformable ProtoPNet~\citep{Donnelly_2022_CVPR}.
While prototypical part-based case-based reasoning has shown to be the most human comprehensible explanation~\citep{nguyen2021effectiveness,kim2023help,jeyakumar2020can}, the number of explanations to be evaluated in its current implementations grows with the number of prototypical parts per class learned by the model.

\paragraph{Contribution}
We propose \ourmod, a transparent neural network for \textbf{s}imilarity-based \textbf{i}nterpretable image \textbf{c}lassifciation, that fulfills all three properties: local, global, and faithful explanations (see Tab.~\ref{tab:properties}).
We inject case-based reasoning into \ourmod's decision-making to provide local and global explanations that reduce the complexity of explanations to be evaluated. 
\ourmod\ can implement any neural network architecture in its feature extractor without limiting their computational capacity.
Specifically, the model extracts support vectors from training images that capture only class-representative features by suppressing negative contributions from other classes.
Classification is performed by computing and summing similarity scores between these support vectors and the latent feature vector of the input image (see Fig.~\ref{fig:fig1}).
Leveraging B-cos networks ensures that both the support vectors and their similarity to test images are directly and faithfully explainable in the input space, enabling an intuitive understanding of the model's reasoning.
In summary, the support samples serve two purposes: (i) they help the model capture the high variability between different classes in the data, and (ii) they provide the user with examples that illustrate which parts of the class-representative training images the feature extractor focuses on.
Our contributions are as follows:
\begin{itemize}
    \item We propose \ourmod, an inherently interpretable network that provides faithful local and global explanations for image classification with either single or multiple labels.
    \item We prove that explanations provided by \ourmod\ fulfill the axioms required to be faithful.
    \item We practically evaluate \ourmod's explanations based on the criteria proposed in the FunnyBirds framework and its associated dataset.
    \item We empirically evaluate \ourmod\ and its explanations on three tasks (fine-grained single-label image classification on Stanford Dogs, pathology prediction on RSNA, multi-label image classification on Pascal VOC), and three architectures with different numbers of parameters (DenseNet121: $\sim$8M, ResNet50: $\sim$26M, a hybrid vision transformer: $\sim$81M).
    \item We demonstrate how the explanations provided by \ourmod\ enable model debugging.
\end{itemize}

\section{Background}\label{sec:ebackground}

\paragraph{Learning Class-Discriminative Support Vectors}

\citet{nwhead_interpret} proposed to classify images with the non-parametric Nadaraya-Watson head and a convolutional neural network $\mathcal{F}_\theta$ with parameters $\theta$ extracting latent vector of dimension $d$.
During training, they randomly sample a number of support vectors $v_k^c$ for every batch per class from the training set $\{(\mathcal{I}_k, y_{k})\}_{k=1}^N$,
$V^c = \left\{ v_k^c = \mathcal{F}_\theta(\mathcal{I}_k)~\vert~ y_k = c \right\}$, with $\Omega(V^c) = N_s$.
This selection is executed for every training batch, thereby directing the optimization of the feature extractor $\mathcal{F}_\theta$ to form distinct, class-specific clusters within the latent feature space.

For classification, a test image $\mathcal{I}$ is first transformed into its latent representation $\mathcal{F}_\theta(\mathcal{I})$. The probability that $\mathcal{I}$ belongs to class $c$ is computed by applying the softmax function to the negative squared L2 distances between $\mathcal{F}_\theta(\mathcal{I})$ and each support vector $v_k^c$:
\begin{align*}
p(y = c \mid \mathcal{I}) = \displaystyle\sum_{k}\frac{\exp\left( -\left\| \mathcal{F}_\theta(\mathcal{I}) - v_k^c \right\|_2^2 \right)}{ \displaystyle\sum_{k'=1}^{N_s} \displaystyle\sum_{c'=1}^C\exp\left( -\left\| \mathcal{F}_\theta(\mathcal{I}) - v_{k'}^{c'} \right\|_2^2 \right)}.
\end{align*}
This classification strategy leverages the learned latent space structure to assign class probabilities based on proximity to the nearest support clusters.

Local explanations show the images of the most influential support samples in terms of softmax probabilities. 
At first glance, these explanations seem intuitive, but the main limitations are two-fold:
(i) Computing probabilities from (inverted) Euclidean distances with softmax:
Regardless of the input (e.g., in an adversarial attack), the model explanations will suggest that at least one support sample is similar with sufficient probability.
Additionally, the softmax probabilities are inapplicable to multi-label classification tasks out of the box since they sum to one.
Thus, the model cannot predict three classes simultaneously under a standard decision threshold of 0.5.
(ii) No pixel-level explanations:
While explanations using similar input images are intuitive for humans, the lack of attribution maps hinders a further understanding of the network's mechanisms and the detection of biased predictions. 

\paragraph{Faithful Explanations with B-cos Networks}

 B-cos networks proposed by~\citet{mboehle} impose weight-input alignment between the layer weights and input by re-formulating the scalar product used by neural network computations, removing biases and standard non-linearities.
This forces the network to focus on the most salient features across the training set and facilitates the summary of each forward pass as an input-dependent linear equation.
Each input feature's contribution can directly be traced by its corresponding weight in the corresponding transformation matrix.

Mathematically, the B-cos transformation computes a modified dot-product, that increases weight-input alignment with an exponent $B$, between an input vector $x$ and a layer weight $w$:
\begin{align*}
    \text{B-cos}(x;w) = \lVert \hat{w} \rVert \lVert x \rVert \lvert \text{cos}(x, \hat{w}) \rvert^B \times \text{sgn}(\text{cos}(x, \hat{w})),
\end{align*}
with $\hat{w} = w / \lVert w \rVert \implies \lVert \hat{w} \rVert = 1$, cos the cosine similarity, sgn the sign function, and $\times$ the real-valued multiplication. While it is bounded by the magnitude of $\Vert x \Vert$, it effectively computes an input-dependent linear equation, in matrix form expressed by $\text{B-cos}(x, \mathbf{W}) = \mathbf{\tilde{W}}(x)x$, with $\mathbf{\tilde{W}}(x) = \lvert \text{cos}(x, \mathbf{\hat{W}})\rvert^{B-1} \odot \mathbf{\hat{W}}$, 
where $\lvert \text{cos}(x, \mathbf{\hat{W}})\rvert^{B-1}$ scales the elements of the fixed weight matrix $\hat{\mathbf{W}}$ ($\odot$ the elementwise multiplication).
A neural network $\mathcal{F}_\theta$ with parameters $\theta$ that computes a series of these equations is called piece-wise linear and its forward pass is summarized by the single input-dependent weight matrix:
\begin{align}\label{eq:summary}
\mathcal{F}_\theta({x}) & = \tilde{\mathbf{W}}_L(a_L)\tilde{\mathbf{W}}_{L-1}(a_{L-1})\dots\tilde{\mathbf{W}}_1(a_1=x)x \nonumber \\ & = \left(\prod_{j=1}^L\tilde{\mathbf{W}_j}(a_j)\right)x = \mathbf{W}_{1 \rightarrow L}(x)x.
\end{align}
Thus, the contribution maps in terms of a pixel location $(m,n)$ in image space for a single forward pass are computed with $\phi^l_j(x)_{(m,n)} = \sum_{ch} \left[\left[\mathbf{W}_{1 \rightarrow l}\right]^T_j \odot x\right]_{(ch,m,n)}$, 
with $l$ any layer and $j$ the index of the row (neuron) in $\mathbf{W}_{1 \rightarrow l}$ (note that this allows to explain any intermediate layer's $l$ neurons $j$).
Since B-cos networks are trained with binary cross-entropy loss, features negatively attributing the class logit cannot increase its probability and are not visualized.

\citet{mboehle} proposed to compute RGBA explanations by normalizing each color channel pair\footnote{B-cos networks require an image encoding of [R, G, B, 1-R, 1-G, 1-B] in the channel dimension to uniquely encode colors and to mitigate favoring of brighter regions.} to sum to 1 to maintain the angle of each color pair and to then scale RGBA values into the range [0..1].
The alpha value (pixel opacity) is computed with the 99.9th percentile $p_{99.9}$ of the resulting weight $w$ as $\text{min}\left(\left[\lVert w_{(m,n)}\rVert_2\right]/\left[p_{99.9}(\lVert w_{(m,n)}\rVert_2)\right], 1\right)$ and smoothed with an average kernel of size nine.

B-cos models offer the notable advantage of generating local explanations from a single forward pass that resembles the network’s transformation of the input image.
However, whether B-cos meets the axioms of faithful explanations remains unclear.
In Appendix~\ref{app:axiomproofs}, we prove that B-cos satisfies these necessary axioms.

\section{Methods}

We propose \ourmod, which combines the strengths of B-cos networks and the Nadaraya-Watson head, to obtain an \emph{inherently interpretable} neural network for similarity-based classification that provides \emph{faithful local} and \emph{global} explanations with a focus on minimizing the number of explanations requiring review.
\ourmod\ achieves this by extracting class-representative support vectors from training images for similarity-based classification and by leveraging the B-cos transform to faithfully explain its decision.
Our implementation is available at~\url{www.github.com/ai-med/SIC}.

\subsection{Architecture}

\begin{figure*}[t]
    \centering
    \includegraphics[width=\textwidth]{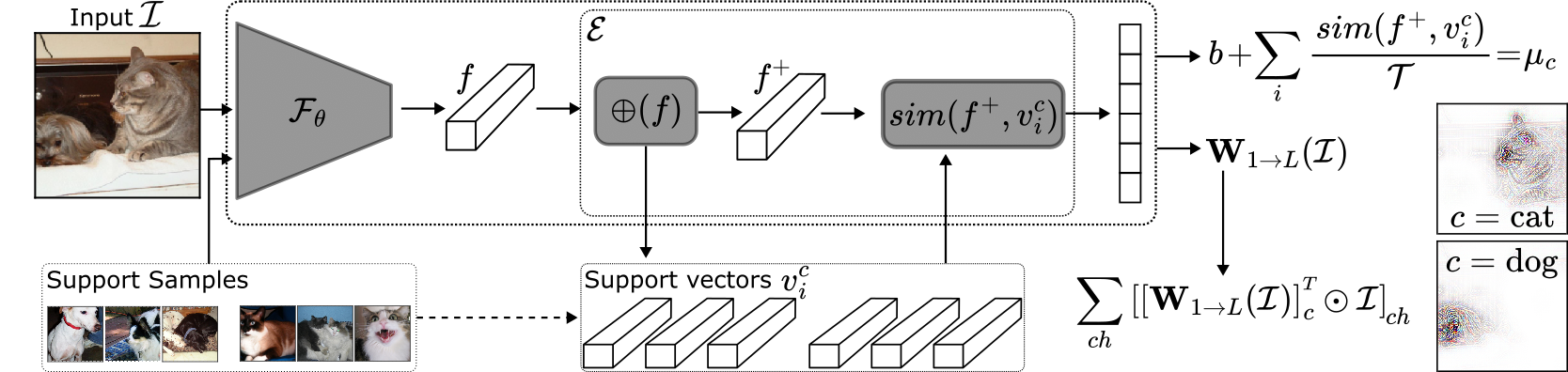}
    \caption{\ourmod\ consists of a feature extractor $\mathcal{F}$ with model weights $\theta$ that extracts latent vectors $f$. The evidence predictor $\mathcal{E}$ converts them into positive-valued vectors $f^+$, which, given other images from the training set, can also serve as support vectors $v_i^c$. Then, it computes the similarity between the input image's latent vector $f$ and support vectors $v_i^c$. Summing the class-specific temperature-normalized similarity scores and adding a bias $b$ yields the class logit $\mu_c$. The B-cos transform facilitates the summary of a forward pass as a single weight-matrix $\mathbf{W}_{1\rightarrow L}(\mathcal{I})$, which is leveraged to compute faithful contribution maps.}
    \label{fig:goat}
\end{figure*}

We visualize \ourmod\ in Fig.~\ref{fig:goat}. It consists of a B-cos feature extractor $\mathcal{F}_\theta$, which transforms an image $\mathcal{I}$ into latent vectors $f = \mathcal{F}_\theta(\mathcal{I}) \in \mathbb{R}^d$, with $\mathcal{I} \in \mathbb{R}^{ch \times H \times W}$, $d$ the latent dimension, $ch$, $H$, and $W$ the image's channels, height, and width, respectively, and $\theta$ the model weights ($\mathcal{F}_\theta$ can be any B-cos neural network architecture).

The evidence predictor $\mathcal{E}$ consists of a function $\oplus: \mathbb{R}^d \rightarrow \mathbb{R}^d_{\geq 0}$ that transforms the real-valued vectors $f$ into non-negative vectors $f^+ = \oplus(f)$, and a similarity measure that computes similarities between support vectors $v_i^c \in F^c = \{f^+_k = \mathcal{F}_\theta(\mathcal{I}_k) ~\vert~ y_k = c\}$ and $f^+$ as $sim(f^+, v_i^c) \in \mathbb{R}$, with $\Omega(F^c) = N_s$ and $i$ denoting the $i$-th support vector of class $c$.
Logits are computed for class $c$ as the sum of its similarity scores 
$\mu_c = b + \sum_{v_i^c} \frac{sim(f^+, v_i^c)}{\mathcal{T}}$, 
where $b$ represents a fixed bias term, $i = 1, \dots, N_s$ indexes the support samples, and $\mathcal{T}$ is a temperature to improve convergence by scaling the logits' magnitude appropriately.

\subsection{Training and Optimization}

Following~\citet{nwhead_interpret} (``closest cluster''), we randomly sample the training set to extract $N_s$ support vectors $v_k^c$ per class to encourage the model to form distinct, class-specific clusters in latent space.
After every epoch, we compute sets of latent vectors $\hat{V}^c = \{f^+_k = \mathcal{F}_\theta(\mathcal{I}_k) ~\vert~ y_k = c \}_{k=1}^N$, which we each cluster using k-means to extract centroids $\gamma_i^c$.
Then, we replace each centroid $\gamma_i^c$ with the closest latent vector to yield support vectors from the same class, defined as
$v_i^c = \arg \min_{f^+_k \mid y_k = c} \Vert f^+_k - \gamma_{i}^c \Vert_2$,
which facilitates case-based reasoning similar to k-nearest-neighbors, as only features descriptive of the class can increase the probability of that class by definition of \ourmod.

We follow~\citet{mboehle_trans} and optimize the model with binary cross-entropy loss, which effectively increases alignment pressure compared to training with standard multi-class cross-entropy loss.

\subsection{Explainability}

\paragraph{Global Explanations}

While global explanations provided by~\ourmod\ are ideal for a developer to debug a model, they also facilitate case-based reasoning, which is most intuitive to understand for humans~\citep{nguyen2021effectiveness,kim2023help,jeyakumar2020can}.

First, we evaluate what is encoded in each support vector $v_i^c$ by computing explanations given the corresponding support image and its extracted feature vector. This feature vector equals $v_i^c$ before its rescaling to the unit norm, simplifying the B-cos transformation to $\text{B-cos}(v_i^c, \hat{v}_i^c) = \lVert v_i^c \rVert$, since $\hat{v}_i^c = v_i^c / \lVert v_i^c \rVert$ and $ \lVert \hat{v}_i^c \rVert = 1$ and $\text{cos}(v_i^c, \hat{v}_i^c)^B = 1$ for any $B$.

We observed negative attributions when computing contribution maps this way during initial experiments.
While only visualizing features with positive class contributions as explanations in RGBA space is reasonable during the classification of an input image, these negative attributions can still serve as a means of debugging a model if carefully evaluated. 
The definition $\text{B-cos}(v_i^c, \hat{v}_i^c) = \lVert v_i^c \rVert$ implies that negative attributions are a means of the model to downscale the range of the B-cos similarity (accordingly, we observed the negative attributions in the corners of the images).
This indicates the temperature scaling $\tau$ was chosen too small, and re-training with a more significant temperature helped to overcome this issue.

Additionally, \ourmod's decision-making is based on a fixed set of latent support vectors.
Hence, we can compute the inter- and intra-class similarities to gain insights into the latent space learned by the feature extractor.

\paragraph{Local Explanations}

Local explanations are threefold:
First, the temperature-scaled similarity scores $\frac{sim(f^+, v_i^c)}{\mathcal{T}}$ (support evidence) provided by the evidence predictor $\mathcal{E}$ quantify the alignment of a test image with a support sample, and their proportion of the log-probability.
Second, since a test image is assigned to a class if it is similar to that class's support sample, 
understanding which parts of the support sample's image are compressed in the support vectors themselves is crucial, i.e., explaining the model mechanisms.
The B-cos properties of \ourmod\ allow us to faithfully compute the RGBA explanations in terms of the similarity measure.
Since each prediction can be summarized by a linear equation, we compute contribution maps in terms of the class logit (test contribution) following the linearity axiom~\citep{axioms} ($\phi = \alpha \phi_1 + \beta \phi_2 \text{ iff } \mathcal{F}(x)= \alpha \mathcal{F}_1(x) + \beta \mathcal{F}_2(x)$).
Therefore, the test sample contributions highlight the image features that aligned with the support vectors.
Third, the global RGBA explanations (support contributions) of the support samples allow the user to check for the intersection of support contributions and test contributions, similar to k-nearest-neighbor.

\paragraph{Theoretic Evaluation}

To thoroughly assess the interpretability of \ourmod, we conduct a theoretical evaluation of its explanations based on the axioms defined by~\citet{axioms}, namely \textit{Completeness, Sensitivity, Implementation Invariance, Dummy, Linearity,} and \textit{Symmetry-Preserving}.
Our proofs (see App.~\ref{app:axiomproofs}), confirm that \ourmod's explanations adhere to these principles, thereby validating the model's capability to provide faithful and comprehensive insights into its decision-making processes.
Thus, \ourmod\ solves the limitations of previous work, which either does not provide pixel-level explanations, like the NW-Head~\citep{nwhead_interpret} or lack global explanations like B-cos neural networks~\cite{mboehle_trans}.

\section{Experiments and Results}

We evaluate \ourmod\ on three tasks with three backbone architectures as feature extractors.
The tasks are 
(i) multi-label image classification on Pascal VOC~\citep{voc}, 
(ii) fine-grained image classification with mutually exclusive labels on Stanford Dogs~\citep{Dogs}, and 
(iii) pathology prediction with mutually exclusive labels on RSNA~\citep{rsna}. 
Additionally, we evaluate the explanations given by \ourmod\ on the FunnyBirds framework~\citep{hesse2023funnybirds}.
The backbone architectures are 
B-cos versions of DenseNet121~\citep{huang2017densely}, ResNet50~\citep{He_2016_CVPR}, and a Hybrid Vision Transformer~\citep{xiao2021early} (VitC), with $\sim$8M, $\sim$26M, and $\sim$81M trainable parameters, respectively.

\begin{figure}[ht]
    \centering
    \includegraphics[width=0.85\linewidth]{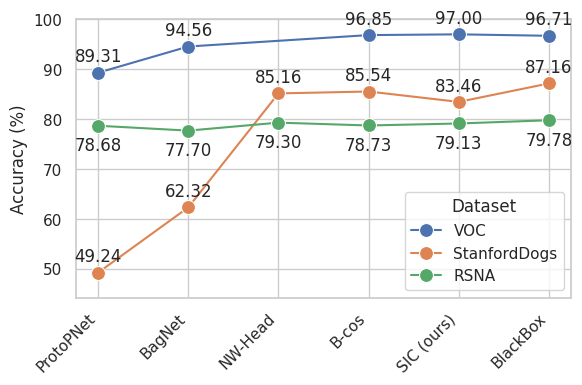}
    \caption{Results in terms of accuracy for ResNet backbones on Pascal VOC and Stanford Dogs, and DenseNet backbones on RSNA. \ourmod\ performs superior to nine out of 14 models. Refer to Tab.~\ref{tab:searchspace}) for exhaustive results.}
    \label{fig:results_acc}
\end{figure}

First, we compare the performance of \ourmod\ in terms of the accuracy (balanced accuracy on RSNA) to black-box neural networks and established inherently interpretable neural networks, namely ProtoPNet~\cite{chen2019looks}, a BagNet17~\cite{brendel2019approximating}, the NW-Head~\cite{nwhead_interpret}, and B-Cos neural networks~\cite{mboehle_trans} in Fig.~\ref{fig:results_acc} and Tab.~\ref{tab:searchspace}.
Compared to black-box versions of the backbones, \ourmod\ performs on par on Pascal VOC and RSNA ($\geq+0.29\%$ and $-0.65\%$ Acc, respectively), and worse on Stanford Dogs ($-3.7\%$ Acc).
Overall, ProtoPNet and BagNets perform worse than all competing methods, and \ourmod\ performs better than nine out of 14 models.

\begin{figure}[!ht]
    \centering
    \begin{subfigure}{\linewidth}
        \centering
        \includegraphics[width=\linewidth]{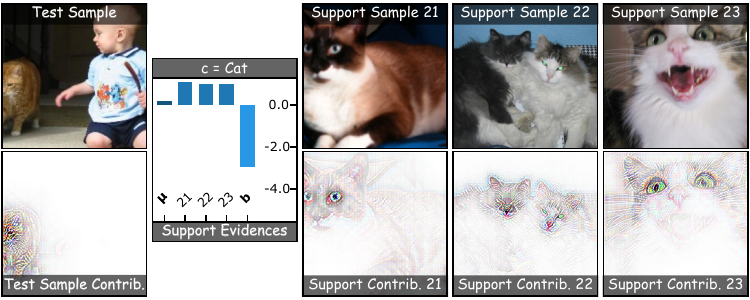}
        \caption{Prediction for the cat class.}
        \label{fig:image1}
    \end{subfigure}
    
    \begin{subfigure}{\linewidth}
        \centering
        \includegraphics[width=\linewidth]{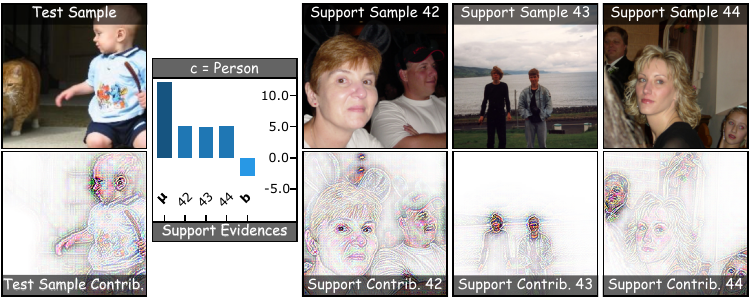}
        \caption{Prediction of the person class.}
        \label{fig:image2}
    \end{subfigure}
    
    \begin{subfigure}{\linewidth}
        \centering
        \includegraphics[width=\linewidth]{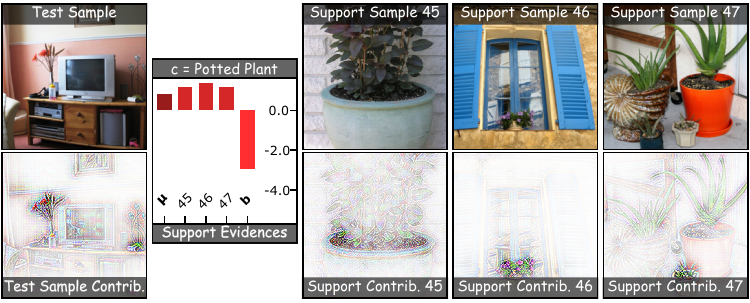}
        \caption{A wrong prediction: The model mistook the flower with potted plants.}
        \label{fig:image3}
    \end{subfigure}
    
    \begin{subfigure}{\linewidth}
        \centering
        \includegraphics[width=\linewidth]{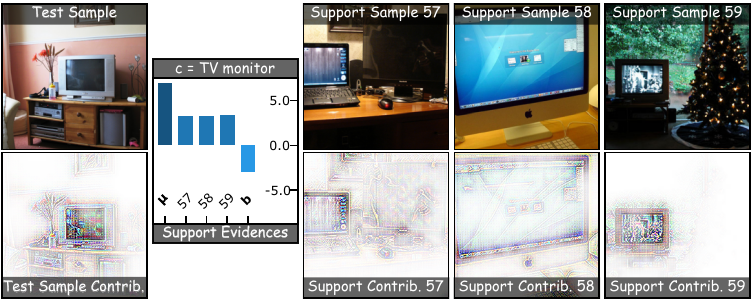}
        \caption{However, the model correctly identified the TV.}
        \label{fig:image4}
    \end{subfigure}
    
    \caption{Correct classifications are depicted with blue support evidence bars, incorrect ones with red. The model trained on Pascal VOC with a DenseNet121 backbone barely classified the test sample in (a) as a cat, because all support images suggest that the model learned to look for two cat eyes. However, the test sample only contains one eye of the cat, and its support contributions suggest that the model is susceptible to detecting two eyes. Computing the contributions for the person class in (b) shows how the model utilizes the other portion of the input features. This shift is also present in the example (c) and (d).}
    \label{fig:all_images}
\end{figure}

\paragraph{Pascal VOC~\citep{voc}}
Fig.~\ref{fig:all_images} illustrates explanations of \ourmod\ for multi-label image classification on Pascal VOC with a DenseNet121 backbone. 
As seen in Figs.~\ref{fig:all_images} (a) and (b), \ourmod\ shifts its focus in the test image depending on the class to predict, i.e., the corresponding column $c$ in $\left[\mathbf{W}_{1\rightarrow L}\right]^T_c$: To predict the cat class, only the area around the cat is present in the contribution map.
When predicting the human class, the complementary area of the image is contained in the contribution map.
The support contributions show that the model is particularly sensitive to the eyes of a cat.
In the test sample in Fig.~\ref{fig:all_images} (a) and (b), only the left half and one eye of the cat are visible, which explains the comparably low support evidences in this particular instance.
Arguably, the support contributions of the human class in Fig.~\ref{fig:all_images} (b) suggest that the model learned to look for human eyes in support sample 42, which shows high contributions from the two lights on the ceiling.
For support samples 43 and 44, the whole heads of the humans are the primary contributions to the support vectors.

Observing the wrong prediction presented in Fig.~\ref{fig:all_images} (c) for the class potted plant, we see that the support contribution map of support sample 45 focuses on the soil in the pot, while the support samples 46 and 47 focus on the actual plants themselves.
We see in the contribution map of the test sample that the model mistakes the flower on the left-hand side and the artificial flowers on the right-hand side with this class.
The model correctly classified the TV/monitor class for the same sample, presented in Fig~\ref{fig:all_images} (d), in which the test contribution map primarily highlights the TV.
Here, the support contributions of sample 57 show that the support vector carries information on all three monitors present in the image.
In the case of support sample 58, the support contribution shows high contributions from the keyboard in addition to the monitor, indicating that the model picked up this bias from the data.
Support contribution 59 focuses on the TV instead of background objects.

We present additional results for other test images and backbones in Figs.~\ref{fig:voc_resnet}-\ref{fig:voc_vitc}.
In many test sample instances (e.g., compare Fig.~\ref{fig:voc_resnet} (a) vs. (b), Fig.~\ref{fig:voc_resnet} (c) vs. (d), Fig.\ref{fig:voc_vitc} (a) vs. (b)), we observe a strong shift in contributions depending on the class prediction.
This demonstrates that the latent vectors of samples with multiple labels efficiently encode features for all classes that need to align with the class support vectors.
In Fig.~\ref{fig:voc_resnet_wrong} (a)-(d), we see that the model predicted four classes, for which, according to the dataset labels, only a single class was actually contained (chair).
However, the explanations look very reasonable. 
Indeed, there is a table in the scene; whether it is a dining table is obviously debatable.  There is also a child in the image, so predicting humans is not unreasonable.
Lastly, the potted plant class logit is very low, and the test contribution map highlights the flower on the table.

For the test sample predicted in Fig.~\ref{fig:voc_vitc} (d) and (e), we see the correct classification of humans with high support evidence scores (note how the model extracted the human for support sample 44 in very dark lighting conditions), and \ourmod\ believed that the martial arts club logo on the person's outfit resembled the logo of a drink.

\paragraph{Stanford Dogs~\citep{Dogs}}

We evaluate \ourmod\ with a ResNet50 backbone on the Stanford dogs dataset.
The contribution maps for the test samples in Figs.~\ref{fig:dogs_resnet_0},~\ref{fig:dogs_resnet_1},and ~\ref{fig:dogs_resnet_2} show that \ourmod\ consistently focuses on the dogs and discards background elements such as humans, especially in Figs.~\ref{fig:dogs_resnet_0} (a) and (b).

The wrong classification of the test sample presented in Fig.~\ref{fig:dogs_resnet_0} (c) shows that primarily the head of the dog contributed to an increased similarity with the support vector of support sample 46, whose support contribution focuses on the head, too.
In contrast, the support contributions of support samples 45 and 47 both indicate that the support vectors contain significant portions of the body of the dog.
Evaluating the explanations of the correct class in Fig.~\ref{fig:dogs_resnet_0} (d) shows that the support sample 33 containing two dogs has high support evidence, as there are two dogs of its class in the test image.
However, the lower support evidence of 34 and 35, which both contain one dog, indicate that the number of instances is encoded into support vector 33.

Considering the support evidence of support sample 162 in Figs.~\ref{fig:dogs_resnet_2} (a) and (b),
its support evidence for the test sample in Fig.~\ref{fig:dogs_resnet_2} (a) is lower compared to the test sample in Fig.~\ref{fig:dogs_resnet_2} (b), which only contains the lower portion of the head like support sample 162.
The support contributions of the other support samples that focus on the eyes and the body show the inverse behavior of the two respective test samples.
However, we observe alike support evidence for all support samples for other classes, e.g., Fig.~\ref{fig:dogs_resnet_1} (a) and (b).
This exemplifies that \ourmod\ is able to capture the distribution of more challenging classes.

\paragraph{RSNA~\citep{rsna}}

We utilize the RSNA dataset to showcase~\ourmod\ on a decision-critical task to differentiate between images with lung opacity and others (healthy and unhealthy without lung opacity, indicated by the presence of bounding boxes) from chest X-ray images.
Here, we want to verify reasonable decision-making by evaluating explanations in terms of bounding boxes. An explanation of a classifier trained to predict lung opacity should focus on an area inside the bounding box, illustrated in Figs.~\ref{fig:rsna_dense_cor} and~\ref{fig:rsna_dense_crit}.

Here, the model has to learn very subtle intra- and inter-class differences.
Evaluating the prototypes in Fig.~\ref{fig:rsna_densenet_protos} shows only positive contributions as desired.
In the case of the three non-occluded support samples on the left, we immediately see that the sample is likely an outlier in that it has some disease without opacity.
The second and third support contributions are primarily highlighted within the lungs and bounding boxes for the respective classes.
Interestingly, the model seems to exclude atypical objects from the healthy support vectors:
Consider the third non-occluded support contribution that does not contain contributions for the dark clamp present in the support sample.
However, the first two support contributions of each class contain most of the input and are, thus, hard to interpret. 

The predictions in Figs.~\ref{fig:rsna_dense_cor} (a)-(d) exemplify how the support evidence varies across the class's support samples and suggest that the increasing number of support samples allowed \ourmod\ to reproduce the class's latent space with a likely high variability.
Additionally, we observe a high degree of intersection between significant test sample contributions and bounding boxes for the occluded class in Figs.~\ref{fig:rsna_dense_cor} (a)-(d) and that the focus of the model is within the lungs for non-occluded samples in Fig.~\ref{fig:rsna_dense_crit}, indicating that the model indeed learned to differentiate occlusion from no-occlusion using medically reasonable regions.

\begin{figure}[t]
    \centering
    \includegraphics[width=\linewidth]{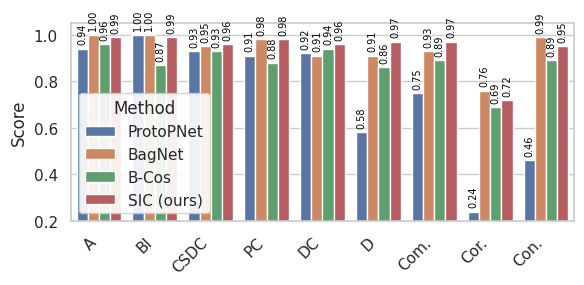}
    \caption{Results on the FunnyBirds Framework~\cite{hesse2023funnybirds}. Note that \textit{Completeness} (Com.) is the mean across Controlled Synthetic Data Check (CSDC), Preservation Check (PC), Deletion Check (DC), and Distractability (D); Accuracy (A), Background Independence (BI), Correctness(Cor.), Contrastivity (Con.). Refer to Tab.~\ref{tab:fbirds_numbers} for full results.}
    \label{fig:results_fb_plot}
\end{figure}

\paragraph{FunnyBirds~\citep{hesse2023funnybirds}}

This framework contains a dataset of synthetic renderings of 50 bird species that are defined by the five concepts beak, wings, feet, eyes, and tail.
Removing or replacing a concept with one of another class enables analysis of explanations in terms of \textit{completeness}, \textit{background independence}, \textit{contrastivitiy}, and \textit{correctness}.

The results, presented in Figs.~\ref{fig:results_fb_plot} and~\ref{fig:fbirds_plot}, and Tab.~\ref{tab:fbirds_numbers}, show that \ourmod\ outperforms ProtoPNet~\cite{chen2019looks} and B-cos~\cite{mboehle} in eight out of nine explainability metrics. 
However, BagNet~\cite{brendel2019approximating} has a lead in terms of \textit{accuracy} (+1\%), \textit{correctness} (+4\%), and \textit{contrastivity} (+4\%).
The explanations for the representative test sample in Fig.~\ref{fig:fbirds_prediction} show that \ourmod\ extracted a diverse range of poses with each support sample offering a distinct view of the relevant concepts. 
Moreover, \ourmod\ demonstrates near-perfect resilience to background noise, evidenced by a \textit{background independence} of 0.99 and the support contributions shown in Fig.~\ref{fig:fbirds_prediction}.
A \textit{contrastivity} of 0.95 further confirms that only class-discriminative features were learned, which aligns with our model design intent.
\textit{Correctness} reflects the assumption that iterative removal of the most important parts from a test image produces a corresponding decline in the classification probability.
\ourmod\ achieves 0.72 \textit{correctness}, which is in line with the scores ($<0.80$) reported by~\citet{hesse2023funnybirds}.
A potential explanation is that the part importance is computed by summing attributions scores within a part without normalization for the part's size.
Finally, a \textit{completeness} of 0.97 suggests that \ourmod\ encoded all class-representative features within its support vectors, as outlined in the support contributions in Fig.~\ref{fig:fbirds_prediction}.

\paragraph{Debugging \ourmod}

While the experiments primarily highlighted the explainability features from the view of a user, we now exemplify how the global and local explanations help to identify flaws in the optimization process. 
Revisiting the model trained on RSNA, we detail how to use the global explanations of \ourmod\ to gain more insights into the model behavior. 
We observe that the support evidence of support sample 0 of the correct classifications in Figs.~\ref{fig:rsna_dense_crit} (a) and (b) is almost absent compared to the wrong classification in Fig.~\ref{fig:rsna_dense_crit} (c).
Considering that support contribution 0 contains large parts of the input, this raises suspicion.
In addition, we see that the only significant evidence for the correct class depicted in Fig.~\ref{fig:rsna_dense_crit} (d) comes from support sample 0, which suffers from contributions across the whole input as well.
As a result, we evaluate the latent space learned by \ourmod\ in Fig.~\ref{fig:tsne_rsna} and find that support vectors 0 and 3, although of different classes, are very close in the t-SNE projection.
We attribute this to a weak latent representation, suggesting that the model has yet to learn a more class-separable latent space.

Next, we evaluate whether the model indeed learned class-representative support vectors (see Fig.~\ref{fig:voc_protos}) by computing the Silhouette score (0.655)~\citep{rousseeuw1987silhouettes} and plotting the inter- and intra-class B-cos similarities of \ourmod\ trained on Pascal VOC with a DenseNet121 backbone in Fig.~\ref{fig:voc_heatmap}.
We observe that the intra-class similarities on the diagonal are indeed higher than those off-diagonal, suggesting good class distinction in latent space.
Additionally, the average distance between some classes is a bit higher, e.g., dining table, chair, and bottle.
This is expected, as images containing a dining table often include a chair and a bottle, requiring the model to extract a latent vector capable of predicting all three classes with sufficient confidence.

\section{Discussion and Conclusion}

Current inherently interpretable models either lack global explanations, fail to explain the internal mechanisms of the deep learning model, or cannot provide explanations that faithfully explain their decision-making.
To this end, we proposed \ourmod\ for similarity-based interpretable classification by providing faithful local, global, and pixel-level explanations. 
Our theoretical analysis proved that our proposed method's explanations satisfy established axioms for explanations, namely ~\textit{Completeness},~\textit{Sensitivity},~\textit{Impementation Invariance},~\textit{Dummy},~\textit{Linearity}, and ~\textit{Symmetry-Preserving}.

We evaluated our method, \ourmod, across three tasks using three distinct backbone architectures, demonstrating that the complex mechanisms learned by neural networks can be decomposed into intuitive, low-complexity explanations.
Our practical evaluation employing the FunnyBirds framework further illustrated that the careful design of \ourmod\ manifests in its explanations.
Notably, our method outperformed established inherently interpretable approaches on the majority of evaluation metrics, while its classification accuracy remained comparable to that of black-box models in two out of three real-world datasets.

Our results suggest that its intuitive global and local explanations may not only facilitate effective model debugging for developers but may also simplify the understanding of its decision-making process for end-users.

While k-means clustering is fast to optimize, \ourmod\ introduces additional overhead by requiring the computation of support vectors, which is a potential limitation in cases with many classes.
However, as demonstrated with the Stanford Dogs dataset comprising 120 classes, \ourmod\ is readily applicable to datasets with hundreds of classes. Moreover, we note that the number of classes in decision-critical fields such as medicine is generally smaller.

In summary, our results demonstrated that embedding interpretability within complex deep neural networks not only achieves competitive performance on challenging benchmarks but also paves the way for more transparent and trustworthy computer vision systems.

\subsubsection*{Acknowledgments}
This research was partially funded by the Deutsche Forschungsgemeinschaft (DFG, German Research Foundation). The authors gratefully acknowledge the computational
and data resources provided by the Leibniz Supercomputing Centre (www.lrz.de).
We thank Sofi\`ene Boutaj for outstanding efforts in experimenting with B-cos models on medical data.

{
    \small
    \bibliographystyle{ieeenat_fullname}
    \bibliography{references}

\begin{thebibliography}{35}
\providecommand{\natexlab}[1]{#1}
\providecommand{\url}[1]{\texttt{#1}}
\expandafter\ifx\csname urlstyle\endcsname\relax
  \providecommand{\doi}[1]{doi: #1}\else
  \providecommand{\doi}{doi: \begingroup \urlstyle{rm}\Url}\fi

\bibitem[Adebayo et~al.(2018)Adebayo, Gilmer, Muelly, Goodfellow, Hardt, and
  Kim]{adebayo2018sanity}
Julius Adebayo, Justin Gilmer, Michael Muelly, Ian Goodfellow, Moritz Hardt,
  and Been Kim.
\newblock Sanity checks for saliency maps.
\newblock \emph{Advances in neural information processing systems}, 31, 2018.

\bibitem[Aditya et~al.(2011)Aditya, Nityananda, Bangpeng, and Li]{Dogs}
Khosla Aditya, Jayadevaprakash Nityananda, Yao Bangpeng, and Fei-Fei Li.
\newblock Novel dataset for fine-grained image categorization.
\newblock In \emph{Proceedings of the First Workshop on Fine-Grained Visual
  Categorization, IEEE Conference on Computer Vision and Pattern Recognition.
  IEEE, Springs, USA}, 2011.

\bibitem[Bach et~al.(2015)Bach, Binder, Montavon, Klauschen, M{\"u}ller, and
  Samek]{bach2015pixel}
Sebastian Bach, Alexander Binder, Gr{\'e}goire Montavon, Frederick Klauschen,
  Klaus-Robert M{\"u}ller, and Wojciech Samek.
\newblock On pixel-wise explanations for non-linear classifier decisions by
  layer-wise relevance propagation.
\newblock \emph{PloS one}, 10\penalty0 (7):\penalty0 e0130140, 2015.

\bibitem[Barnett et~al.(2021)Barnett, Schwartz, Tao, Chen, Ren, Lo, and
  Rudin]{barnett2021interpretable}
Alina~Jade Barnett, Fides~Regina Schwartz, Chaofan Tao, Chaofan Chen, Yinhao
  Ren, Joseph~Y Lo, and Cynthia Rudin.
\newblock Interpretable mammographic image classification using case-based
  reasoning and deep learning.
\newblock \emph{arXiv preprint arXiv:2107.05605}, 2021.

\bibitem[B\"ohle et~al.(2022)B\"ohle, Fritz, and Schiele]{mboehle}
Moritz B\"ohle, Mario Fritz, and Bernt Schiele.
\newblock B-cos networks: Alignment is all we need for interpretability.
\newblock In \emph{Proceedings of the IEEE/CVF Conference on Computer Vision
  and Pattern Recognition (CVPR)}, pages 10329--10338, 2022.

\bibitem[Brendel and Bethge(2019)]{brendel2019approximating}
Wieland Brendel and Matthias Bethge.
\newblock Approximating cnns with bag-of-local-features models works
  surprisingly well on imagenet.
\newblock \emph{arXiv preprint arXiv:1904.00760}, 2019.

\bibitem[Böhle et~al.(2024)Böhle, Singh, Fritz, and Schiele]{mboehle_trans}
Moritz Böhle, Navdeeppal Singh, Mario Fritz, and Bernt Schiele.
\newblock B-cos alignment for inherently interpretable cnns and vision
  transformers.
\newblock \emph{IEEE Transactions on Pattern Analysis and Machine
  Intelligence}, 46\penalty0 (6):\penalty0 4504--4518, 2024.

\bibitem[Chen et~al.(2019)Chen, Li, Tao, Barnett, Rudin, and Su]{chen2019looks}
Chaofan Chen, Oscar Li, Daniel Tao, Alina Barnett, Cynthia Rudin, and
  Jonathan~K Su.
\newblock This looks like that: deep learning for interpretable image
  recognition.
\newblock \emph{Advances in neural information processing systems}, 32, 2019.

\bibitem[Donnelly et~al.(2022)Donnelly, Barnett, and Chen]{Donnelly_2022_CVPR}
Jon Donnelly, Alina~Jade Barnett, and Chaofan Chen.
\newblock Deformable protopnet: An interpretable image classifier using
  deformable prototypes.
\newblock In \emph{Proceedings of the IEEE/CVF Conference on Computer Vision
  and Pattern Recognition (CVPR)}, pages 10265--10275, 2022.

\bibitem[Everingham et~al.(2009)Everingham, Gool, Williams, Winn, and
  Zisserman]{voc}
M. Everingham, L. Gool, C.~K. Williams, J. Winn, and Andrew Zisserman.
\newblock The pascal visual object classes (voc) challenge.
\newblock \emph{International Journal of Computer Vision}, 88:\penalty0
  303--338, 2009.

\bibitem[He et~al.(2016)He, Zhang, Ren, and Sun]{He_2016_CVPR}
Kaiming He, Xiangyu Zhang, Shaoqing Ren, and Jian Sun.
\newblock Deep residual learning for image recognition.
\newblock In \emph{Proceedings of the IEEE Conference on Computer Vision and
  Pattern Recognition (CVPR)}, 2016.

\bibitem[Hesse et~al.(2023)Hesse, Schaub-Meyer, and Roth]{hesse2023funnybirds}
Robin Hesse, Simone Schaub-Meyer, and Stefan Roth.
\newblock Funnybirds: A synthetic vision dataset for a part-based analysis of
  explainable ai methods.
\newblock In \emph{Proceedings of the IEEE/CVF International Conference on
  Computer Vision}, pages 3981--3991, 2023.

\bibitem[Hoffmann et~al.(2021)Hoffmann, Fanconi, Rade, and
  Kohler]{hoffmann2021looks}
Adrian Hoffmann, Claudio Fanconi, Rahul Rade, and Jonas Kohler.
\newblock This looks like that... does it? shortcomings of latent space
  prototype interpretability in deep networks.
\newblock \emph{arXiv preprint arXiv:2105.02968}, 2021.

\bibitem[Huang et~al.(2017)Huang, Liu, Van Der~Maaten, and
  Weinberger]{huang2017densely}
Gao Huang, Zhuang Liu, Laurens Van Der~Maaten, and Kilian~Q Weinberger.
\newblock Densely connected convolutional networks.
\newblock In \emph{Proceedings of the IEEE conference on computer vision and
  pattern recognition}, pages 4700--4708, 2017.

\bibitem[Jeyakumar et~al.(2020)Jeyakumar, Noor, Cheng, Garcia, and
  Srivastava]{jeyakumar2020can}
Jeya~Vikranth Jeyakumar, Joseph Noor, Yu-Hsi Cheng, Luis Garcia, and Mani
  Srivastava.
\newblock How can i explain this to you? an empirical study of deep neural
  network explanation methods.
\newblock \emph{Advances in Neural Information Processing Systems},
  33:\penalty0 4211--4222, 2020.

\bibitem[Kim et~al.(2021)Kim, Kim, Seo, and Yoon]{kim2021xprotonet}
Eunji Kim, Siwon Kim, Minji Seo, and Sungroh Yoon.
\newblock Xprotonet: diagnosis in chest radiography with global and local
  explanations.
\newblock In \emph{Proceedings of the IEEE/CVF conference on computer vision
  and pattern recognition}, pages 15719--15728, 2021.

\bibitem[Kim et~al.(2023)Kim, Watkins, Russakovsky, Fong, and
  Monroy-Hern{\'a}ndez]{kim2023help}
Sunnie~SY Kim, Elizabeth~Anne Watkins, Olga Russakovsky, Ruth Fong, and
  Andr{\'e}s Monroy-Hern{\'a}ndez.
\newblock "help me help the ai": Understanding how explainability can support
  human-ai interaction.
\newblock In \emph{Proceedings of the 2023 CHI Conference on Human Factors in
  Computing Systems}, pages 1--17, 2023.

\bibitem[Koh et~al.(2020)Koh, Nguyen, Tang, Mussmann, Pierson, Kim, and
  Liang]{CBM}
Pang~Wei Koh, Thao Nguyen, Yew~Siang Tang, Stephen Mussmann, Emma Pierson, Been
  Kim, and Percy Liang.
\newblock Concept bottleneck models.
\newblock In \emph{International conference on machine learning}, pages
  5338--5348. PMLR, 2020.

\bibitem[Lundberg and Lee(2017)]{Lundberg_2017}
Scott~M Lundberg and Su-In Lee.
\newblock A unified approach to interpreting model predictions.
\newblock \emph{Advances in neural information processing systems}, 30, 2017.

\bibitem[Nauta et~al.(2021)Nauta, van Bree, and Seifert]{nauta2021prototree}
Meike Nauta, Ron van Bree, and Christin Seifert.
\newblock Neural prototype trees for interpretable fine-grained image
  recognition.
\newblock In \emph{Proceedings of the IEEE/CVF Conference on Computer Vision
  and Pattern Recognition (CVPR)}, pages 14933--14943, 2021.

\bibitem[Nauta et~al.(2023)Nauta, Schl{\"o}tterer, Van~Keulen, and
  Seifert]{nauta2023pip}
Meike Nauta, J{\"o}rg Schl{\"o}tterer, Maurice Van~Keulen, and Christin
  Seifert.
\newblock Pip-net: Patch-based intuitive prototypes for interpretable image
  classification.
\newblock In \emph{Proceedings of the IEEE/CVF Conference on Computer Vision
  and Pattern Recognition}, pages 2744--2753, 2023.

\bibitem[Nguyen et~al.(2021)Nguyen, Kim, and Nguyen]{nguyen2021effectiveness}
Giang Nguyen, Daeyoung Kim, and Anh Nguyen.
\newblock The effectiveness of feature attribution methods and its correlation
  with automatic evaluation scores.
\newblock \emph{Advances in Neural Information Processing Systems},
  34:\penalty0 26422--26436, 2021.

\bibitem[Petsiuk(2018)]{petsiuk2018rise}
V Petsiuk.
\newblock Rise: Randomized input sampling for explanation of black-box models.
\newblock \emph{arXiv preprint arXiv:1806.07421}, 2018.

\bibitem[Ribeiro et~al.(2016)Ribeiro, Singh, and Guestrin]{ribeiro2016should}
Marco~Tulio Ribeiro, Sameer Singh, and Carlos Guestrin.
\newblock " why should i trust you?" explaining the predictions of any
  classifier.
\newblock In \emph{Proceedings of the 22nd ACM SIGKDD international conference
  on knowledge discovery and data mining}, pages 1135--1144, 2016.

\bibitem[Rousseeuw(1987)]{rousseeuw1987silhouettes}
Peter~J Rousseeuw.
\newblock Silhouettes: a graphical aid to the interpretation and validation of
  cluster analysis.
\newblock \emph{Journal of computational and applied mathematics}, 20:\penalty0
  53--65, 1987.

\bibitem[Rudin(2019)]{Rudin_2019}
Cynthia Rudin.
\newblock Stop explaining black box machine learning models for high stakes
  decisions and use interpretable models instead.
\newblock \emph{Nat Mach Intell}, 1\penalty0 (5):\penalty0 206--215, 2019.

\bibitem[Sacha et~al.(2024)Sacha, Jura, Rymarczyk, Struski, Tabor, and
  Zieli{\'n}ski]{sacha2024interpretability}
Miko{\l}aj Sacha, Bartosz Jura, Dawid Rymarczyk, {\L}ukasz Struski, Jacek
  Tabor, and Bartosz Zieli{\'n}ski.
\newblock Interpretability benchmark for evaluating spatial misalignment of
  prototypical parts explanations.
\newblock In \emph{Proceedings of the AAAI Conference on Artificial
  Intelligence}, pages 21563--21573, 2024.

\bibitem[Selvaraju et~al.(2017)Selvaraju, Cogswell, Das, Vedantam, Parikh, and
  Batra]{selvaraju2017grad}
Ramprasaath~R Selvaraju, Michael Cogswell, Abhishek Das, Ramakrishna Vedantam,
  Devi Parikh, and Dhruv Batra.
\newblock Grad-cam: Visual explanations from deep networks via gradient-based
  localization.
\newblock In \emph{Proceedings of the IEEE international conference on computer
  vision}, pages 618--626, 2017.

\bibitem[Shih et~al.(2019)Shih, Wu, Halabi, Kohli, Prevedello, Cook, Sharma,
  Amorosa, Arteaga, Galperin-Aizenberg, et~al.]{rsna}
George Shih, Carol~C Wu, Safwan~S Halabi, Marc~D Kohli, Luciano~M Prevedello,
  Tessa~S Cook, Arjun Sharma, Judith~K Amorosa, Veronica Arteaga, Maya
  Galperin-Aizenberg, et~al.
\newblock Augmenting the national institutes of health chest radiograph dataset
  with expert annotations of possible pneumonia.
\newblock \emph{Radiology: Artificial Intelligence}, 1\penalty0 (1):\penalty0
  e180041, 2019.

\bibitem[Sundararajan et~al.(2017)Sundararajan, Taly, and Yan]{axioms}
Mukund Sundararajan, Ankur Taly, and Qiqi Yan.
\newblock Axiomatic attribution for deep networks.
\newblock In \emph{International conference on machine learning}, pages
  3319--3328. PMLR, 2017.

\bibitem[Wang and Sabuncu(2022)]{nwhead_interpret}
Alan~Q Wang and Mert~R Sabuncu.
\newblock A flexible nadaraya-watson head can offer explainable and calibrated
  classification.
\newblock \emph{arXiv preprint arXiv:2212.03411}, 2022.

\bibitem[Welinder et~al.(2010)Welinder, Branson, Mita, Wah, Schroff, Belongie,
  and Perona]{welinder2010caltech}
Peter Welinder, Steve Branson, Takeshi Mita, Catherine Wah, Florian Schroff,
  Serge Belongie, and Pietro Perona.
\newblock Caltech-ucsd birds 200.
\newblock 2010.

\bibitem[Wolf et~al.(2023)Wolf, P{\"o}lsterl, and Wachinger]{Wolf_2023}
Tom~Nuno Wolf, Sebastian P{\"o}lsterl, and Christian Wachinger.
\newblock Don't panic: Prototypical additive neural network for interpretable
  classification of alzheimer's disease.
\newblock In \emph{Information Processing in Medical Imaging--28th
  International Conference, IPMI 2023, San Carlos de Bariloche, Argentina, June
  18–23, 2023, Proceedings}. Springer, 2023.

\bibitem[Wolf et~al.(2024)Wolf, Bongratz, Rickmann, P{\"o}lsterl, and
  Wachinger]{wolf2024keep}
Tom~Nuno Wolf, Fabian Bongratz, Anne-Marie Rickmann, Sebastian P{\"o}lsterl,
  and Christian Wachinger.
\newblock Keep the faith: Faithful explanations in convolutional neural
  networks for case-based reasoning.
\newblock In \emph{Proceedings of the AAAI Conference on Artificial
  Intelligence}, pages 5921--5929, 2024.

\bibitem[Xiao et~al.(2021)Xiao, Singh, Mintun, Darrell, Doll{\'a}r, and
  Girshick]{xiao2021early}
Tete Xiao, Mannat Singh, Eric Mintun, Trevor Darrell, Piotr Doll{\'a}r, and
  Ross Girshick.
\newblock Early convolutions help transformers see better.
\newblock \emph{Advances in neural information processing systems},
  34:\penalty0 30392--30400, 2021.

\end{thebibliography}
}

\clearpage

\appendix

\section{Appendix}
\renewcommand{\thefigure}{A.\arabic{figure}}
\renewcommand{\thetable}{A.\arabic{table}}
\subsection{On the Faithfulness of B-cos Explanations}\label{app:axiomproofs}

B-cos modles are piece-wise linear models that allows to summarize each input as an input-dependent linear transform $f({x}) = \mathbf{W}_{1 \rightarrow L}(x)x$, derived in Eq.~\ref{eq:summary}.
Explanations are computed in terms of the weight matrix $\mathbf{W}_{1 \rightarrow L}(x)$ by multiplying each feature with its matrix weight $\phi(x)_{i} = [\mathbf{W}_{1 \rightarrow L}]^T_j \odot x]_{(ch,i,j)}$.
To show that computing explanations for a forward pass is faithful (i.e., satisfies below six axioms introduced by~\citet{axioms}), we need to show that they hold considering the transformation matrix $\mathbf{W}_{1 \rightarrow L}(x)$.
While $\mathbf{W}_{1 \rightarrow L}(x)$ is indeed input-dependent, it is effectively fixed during the computation of the explanation for its specific input $x$.
This is because the explanations $\phi(f, X)$ aim to attribute the output $f(x)$ to the input features of $x$ using the weights at that point. 
Therefore, we assume a fixed $\mathbf{W}_{1 \rightarrow L}(x)$ when explaining the forward pass of an input $x$, and acknowledge that while $\mathbf{W}_{1 \rightarrow L}(x)$ varies across different inputs, it remains constant within the context of computing local explanations $\phi(f, x)$ for a particular $x$.
In addition, we consider the case of single-class prediction, where the weight matrix simplifies to a single-column form ($\mathbf{W}_{1 \rightarrow L}(X) \in \mathbb{R}$) for in input vector $X \in \mathbb{R}$.
This allows us to focus on the core properties of the explanation mechanism without loss of generality since explaining the prediction of a class in the multiclass setting is analogous to explaining its corresponding column vector of the matrix $\mathbf{W}_{1 \rightarrow L}(x)$.

We abbreviate $\mathbf{W}_{1 \rightarrow L}(x)$ with $W(X)$ for ease of notation in the following.
Let ~\ourmod\ be defined as $f(X) = W(X) \cdot X$ and its contributions similarily by $\phi(f, X) = W(X) \odot X = (W(X)_1 \times X_1, \dots, W(X)_N \times X_N)$ (remark that $\cdot$ denotes the scalar product and $\odot$ the element-wise multiplication, $\times$ the multiplication of real numbers).

Now, we reformulate $f(X)$:
\begin{align}\label{eq:reformulate}
f(X) = W(X) \cdot X = \sum_{i=1}^N W(X)_i \times X_i = \sum_{i=1}^N \phi(f, X)_i.
\end{align}
\paragraph{Completeness:}

\textit{The sum of feature attributions should add up to the model output.}

This follows trivially from the reformulation in Eq.\ref{eq:reformulate}.

\paragraph{Sensitivity:}

\textit{If changing only one feature's value changes the prediction of the model, this feature's attribution should be non-zero.}

We reformulate: $f(X) = \sum_{j=1}^N \phi(f, X)_j = \phi(f, X)_i + \sum_{j\neq i}^N \phi(f, X)_j$.

Modifying $X$ only in $i$, denoted by $\hat{X}$, and $f(X) \neq f(\hat{X})$ yields:

$f(X) = \phi(f, X)_i + \sum_{j\neq i}^N \phi(f, X)_j \neq f(\hat{X}) = \phi(f, \hat{X})_i + \sum_{j\neq i}^N \phi(f, X)_j$.

and allows to substitute $\sum_{j\neq i}^N \phi(f, X)_j$.

$\implies \phi(f, X)_i \neq \phi(f, \hat{X})_i$, which means that at least one of the two attributions in non-zero.

\paragraph{Implementation Invariance}

\textit{If two models are functionally equivalent, i.e. the outputs are equal for all inputs, despite having different implementations, their contributions should always be identical.}

Let $f_1, f_2$ be implementations.

Then: $f_1 = W_1(X) \cdot X$, $f_2 = W_2(X) \cdot X, ~ \forall X \implies W_1(X) = W_2(X), ~ \forall X$.

It follows that $W_1(X) \odot X = W_2(X) \odot X, ~ \forall X \implies \phi(f_1, X) = \phi(f_2, X), ~ \forall X$.

\paragraph{Dummy}

\textit{If a model $f$ does not depend on some feature $X_i$, its attribution should always be zero.}

If $f(X)$ is does not depend on some feature $X_i$, it follows that $W(X)_i \times X_i = 0, ~\forall X_i \in \mathbb{R}$

$\implies W(X)_i = 0 ~\forall, X_i \in \mathbb{R}$

$\implies \phi(f, X)_i = 0, ~\forall X_i \in \mathbb{R}$.

\paragraph{Linearity}

\textit{If the output of a model is a linear combination of two models, the attribution of the combined model should be the weighted sum of the contributions of the original models}.

We need to show that $f(X) = \alpha f_1(X) + \beta f_2(X) \implies \phi(f, X) = \alpha\phi(f_1, X) + \beta\phi(f_2, X)$.

$f(X) = \alpha f_1(X) + \beta f_2(X) = \alpha(\sum_{i=1}^N W_1(X)_i \times X_i) + \beta(\sum_{i=1}^N W_2(X)_i \times X_i) = \alpha(\sum_{i=1}^N \phi(f_1, X)_i) + \beta(\sum_{i=1}^N \phi(f_2, X)_i) = \alpha\phi(f_1, X) + \beta\phi(f_2, X)$.

\paragraph{Symmetry-Preserving}

\textit{If swapping two features does not change the model output for all possible values, they should have identical attributions}.

Let $X_i, X_j$ be two features.

If swapping the two does not change the model output for all possible values, it follows that

$W(X)_i \times X_i = W(X)_j \times X_j, ~\forall X_i, X_j \in \mathbb{R}$

$\implies \phi(f, X)_i = \phi(f, X)_j, ~\forall X_i, X_j \in \mathbb{R}$.

\subsection{On the Faithfulness of support vectors}
On a theoretical level, the global explanations are the explanations of an intermediate B-cos neuron.
Thus, they satisfy the above axioms as well and can be interpreted as the information encoded in the respective support vector.

When providing local explanations, we view support vectors as the weights of a B-cos linear transform.
Hence, computing the explanation of the prediction of the class in terms of the output yields the input features compressed in any of the support vectors;
and the log-probability scores of each support vector indicate to which extend each was found.

\paragraph{Justification for ReLU in \ourmod}

In contrast to other functions that could be used to implement $\oplus$, the ReLU activation does not alter the range of the input to potentially huge or tiny numbers like, e.g. the exponential.
However, ReLU activations are commonly scrutinized for their role in setting negative activations to zero, which can cause deep neural networks to interpret this suppression as the absence of a feature.
Thus, many XAI methods (e.g. gradient-based) yield misleading attribution maps, as they cannot account for this type of contribution.
However, this phenomenon is tightly controlled in \ourmod, where the sole operation following the ReLU activation is a similarity computation between the latent vectors of two images.
As a result, the absence of a feature does not, by design, affect the probability of class logits, rendering it valid to disregard any input and its explanation from neurons corresponding to such features.

\subsection{Implementation and Training Details}

\paragraph{Optimization}
All models were optimized using the AdamW optimizer with fixed hyperparameters: weight decay set to 0.0, betas configured to (0.9, 0.999), and an epsilon value of 1e-08.
For FunnyBirds, we selected the default parameters of the framework: a learning rate of 0.001, and weight decay of 0.0, trained for 100 epochs with a temperature of 30 and a batch size of 8.
For all other experiments, we conducted an extensive grid search over the hyperparameter space (see Tab.~\ref{tab:searchspace}) to identify the optimal settings based on accuracy.
The training employed a custom learning rate scheduling policy, which began with a linear warm-up phase over the first two epochs, increasing the learning rate from 10\% to the initial learning rate.
This was followed by maintaining the initial learning rate until 50\% of the total training iterations were completed.
Subsequently, the learning rate was decayed by a factor of 0.5 at every subsequent 10\% milestone of the total iterations.
Batch sizes were set to 32 for all datasets except RSNA, which utilized a batch size of 64. The number of training epochs was standardized to 50 across all datasets, with RSNA trained for 100 epochs. Pretrained feature extractor weights were sourced from torchvision~\url{https://pytorch.org/vision/stable/index.html} and the official B-cos repository~\url{https://github.com/B-cos/B-cos-v2} for all experiments except those involving RSNA.
We train ProtoPNet with the default training recipe provided on the official GitHub implementation (\url{https://github.com/cfchen-duke/ProtoPNet}) and adapted the code for VOC for multi-label classification.
Note that the NW-Head is not applicable to multi-label classification tasks as it relies on softmax for normalization, which can not be adapted with minor modifications as done with ProtoPNet. 

Model-specific parameters included setting B-cos $B=2$ and using three support samples.
We train the NW-Head in the replace cluster mode.
For \ourmod, the grid search explored learning rates of 0.01, 0.003, 0.001, and 0.0003 for the RSNA dataset, and 0.0022, 0.001, 0.00022, 0.0001, 0.000022, and 0.00001 for other datasets. Temperature parameters tested were 10, 30, and 50 for RSNA, and 10 and 30 for the remaining datasets.
For BlackBox models, the grid search included the same learning rates as \ourmod\ and additionally varied the weight decay parameter, evaluating values of 0.0 and 0.0001.
The best-performing models were selected based on their accuracy metrics.

\begin{table*}[ht]
    \centering
    \begin{tabular}{llccccc}
        \toprule
        \textbf{Method} & \textbf{Dataset} & \textbf{Backbone} & \textbf{Best LR} & \textbf{Best WD} & \textbf{Best Temp.} & \textbf{Acc (\%)} \\
        \midrule
        \multirow{5}{*}{BlackBox} & VOC & DenseNet121       & $2.2 \times 10^{-5}$ & 0.0   & -   & 96.42 \\
                                   & VOC & ResNet50          & $1.0 \times 10^{-5}$ & 0.0   & -   & 96.71 \\
                                   & VOC & VitC              & $1.0 \times 10^{-4}$ & 0.0   & -   & 96.35 \\
                                   & StanfordDogs & ResNet50 & $1.0 \times 10^{-5}$ & 0.0   & -   & 87.16 \\
                                   & RSNA & DenseNet121      & $3.0 \times 10^{-4}$ & $1.0 \times 10^{-4}$ & -   & 79.78 \\
        \midrule
                                 \multirow{2}{*}{NW-Head~\cite{nwhead_interpret}}  & StanfordDogs & ResNet50 &  $1.0 \times 10^{-5}$ & $0.0$     & -  & 85.16 \\
                                   & RSNA & DenseNet121      &  $3.0 \times 10^{-4}$ & $1.0 \times 10^{-4}$     & -  & 79.30 \\
        \midrule
        \multirow{4}{*}{ProtoPNet~\cite{chen2019looks}}   & VOC & DenseNet121       &  -                            & -     & -  & 91.19 \\
                                   & VOC & ResNet50       &  -           & -     & -  & 89.31 \\
                                   & StanfordDogs & ResNet50 & - & -     & -  & 49.24 \\      & StanfordDogs & ResNet50* & - & -     & -  & 74.70 \\
                                   & RSNA & DenseNet121      & - & -     & -  & 78.68 \\
        \midrule
        \multirow{3}{*}{BagNet~\cite{brendel2019approximating}}   & VOC & BagNet17       & $2.2 \times 10^{-5}$                             & $1.0 \times 10^{-4}$     & -  & 94.56 \\
                                   & StanfordDogs & BagNet17 &  $2.2\times 10^{-5}$ & 0.0     & - & 62.32 \\
                                   & RSNA & BagNet17      & $3.0 \times 10^{-4}$ & $1.0 \times 10^{-4}$     & -  & 77.70 \\
        \midrule
        \multirow{5}{*}{B-cos~\cite{mboehle}}   & VOC & DenseNet121       &  $1.0 \times 10^{-4}$                            & -     & -  & 96.48 \\
                                   & VOC & ResNet50          & $1.0 \times 10^{-4}$ & -     & -  & 96.85 \\
                                   & VOC & VitC              & $2.2 \times 10^{-5}$ & -     & -  & 96.79 \\
                                   & StanfordDogs & ResNet50 &  $1.0 \times 10^{-4}$ & -    & -  & 85.54 \\
                                   & RSNA & DenseNet121      &  $1.0 \times 10^{-2}$   & -  & -  & 78.73 \\
        \midrule
        \multirow{5}{*}{\ourmod}   & VOC & DenseNet121       & $2.2 \times 10^{-4}$ & -     & 30  & 96.73 \\
                                   & VOC & ResNet50          & $1.0 \times 10^{-4}$ & -     & 30  & 97.00 \\
                                   & VOC & VitC              & $2.2 \times 10^{-5}$ & -     & 10  & 96.72 \\
                                   & StanfordDogs & ResNet50 & $1.0 \times 10^{-4}$ & -     & 10  & 83.46 \\
                                   & RSNA & DenseNet121      & $3.0 \times 10^{-4}$ & -     & 50  & 79.13 \\
        \bottomrule
    \end{tabular}
    \caption{Best configurations and performance of each model.\\ $*$: optimized with training recipe published in~\cite{Donnelly_2022_CVPR} instead of~\cite{chen2019looks}.}
    \label{tab:searchspace}
\end{table*}

\paragraph{Architecture}
All backbones are extended by a linear projection layer to extract feature vectors $\in \mathbb{R}^d=128$.
The $\oplus$ function is implemented as a ReLU activation in the evidence predictor $\mathcal{E}$, and the similarity function ``sim'' is implemented by a B-cos linear layer.
In multi-label classification settings, only training samples with a single class label are considered for support labels.

\paragraph{Datasets, Data Pre-Processing and Augmentation}
In our study, we adhered to standard dataset splits to ensure consistency and comparability with existing research.
Specifically, for Stanford Dogs, and Pascal VOC, we utilized the official training and testing partitions as provided.
Regarding the RSNA dataset, we employed the training set from Stage 2 of the RSNA challenge and further subdivided it by randomly sampling 25\% of the training data to form a separate test set, ensuring that the splits were stratified by class label.
Additionally, we verified that the distributions across sex remained consistent following the splitting process, as illustrated in Tab.~\ref{tab:rsna_ds_stats}.

\begin{table}[ht]
    \centering
    \begin{tabular}{llcr}
        \toprule
        \textbf{Split} & \textbf{Sex} & \textbf{Target} & \textbf{Num. Samples} \\
        \midrule
        \multirow{4}{*}{Train} 
            & F & No Opacity & 6,770 \\
            & F & Opacity     & 1,873 \\
            & M & No Opacity & 8,734 \\
            & M & Opacity     & 2,636 \\
        \midrule
        \multirow{4}{*}{Test} 
            & F & No Opacity & 2,246 \\
            & F & Opacity     & 629 \\
            & M & No Opacity & 2,922 \\
            & M & Opacity    & 874 \\
        \bottomrule
    \end{tabular}
    \caption{Dataset statistics of the RSNA dataset.}
    \label{tab:rsna_ds_stats}
\end{table}

For non-Bcos models, we applied standard normalization using the mean and standard deviation calculated from the training set. Specifically, images were standardized with mean values $[0.485, 0.456, 0.406]$ and standard deviations $[0.229, 0.224, 0.225]$.

Data pre-processing for the Pascal VOC and Stanford Dogs datasets included a \texttt{RandomResizedCrop} of torchvision to a target size of 224 pixels using bilinear interpolation. We implemented a \texttt{RandomHorizontalFlip} with a probability of 0.5. For non-Bcos models, standardization was performed as mentioned above. Additionally, we employed \texttt{RandomErasing} with a probability of 0.5 and a scale range of 2\% to 33\% of the image area.

We follow the FunnyBirds framework and use the default data processing pipeline and compute the metrics with the unsmoothed explanations $\phi$, i.e. the alpha channel of RGBA explanations.

For the RSNA dataset, intensity rescaling was conducted to map the maximum Hounsfield Units to a range between 0 and 1. Data augmentation techniques included a \texttt{RandomHorizontalFlip} with a probability of 0.5 and a \texttt{RandomAffine} transformation. The \texttt{RandomAffine} parameters consisted of rotations up to 45 degrees, translations of \(\pm15\%\) in each direction, and scaling factors ranging from 0.85 to 1.15. We also applied \texttt{RandomErasing} with a probability of 0.5, using patch sizes between 5\% and 20\% of the image area.

\subsubsection{FunnyBirds - ResNet50}

\begin{figure}[hb]
    \centering
    \includegraphics[width=0.47\linewidth]{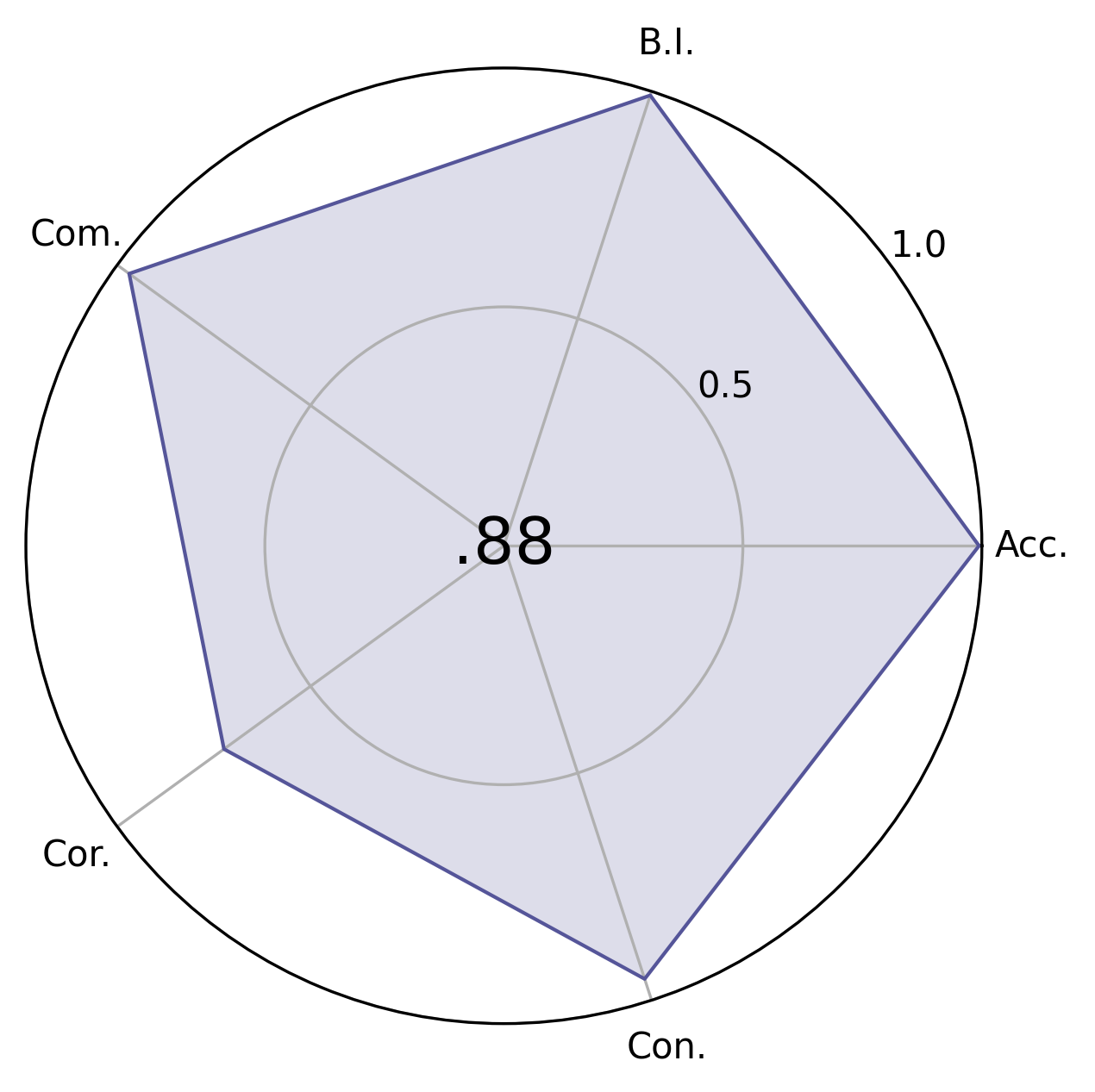}
    \caption{FunnyBirds~\citep{hesse2023funnybirds} result of main metrics Completeness (Com.), Background Independence (B.I.), Accuracy, Contrastivity (Con.), Correctness (Cor.).}
    \label{fig:fbirds_plot}
\end{figure}

\begin{table*}[ht]
    \centering
    \begin{tabular}{lccccccccccc}
        \hline
        \textbf{Method} & \textbf{A} & \textbf{BI} & \textbf{CSDC} & \textbf{PC} & \textbf{DC} & \textbf{D} & \textbf{SD} & \textbf{TS} & \textbf{Com.} & \textbf{Cor.} & \textbf{Con.} \\
        \hline
        ProtoPNet~\cite{chen2019looks} & 0.94 & \textbf{1.00} & 0.93 & 0.91 & 0.92 & 0.58 & 0.24 & 0.46 & 0.75 & 0.24 & 0.46 \\
        BagNet~\citep{brendel2019approximating} & \textbf{1.00} & \textbf{1.00} & 0.95 & \textbf{0.98} & 0.91 & 0.91 & \textbf{0.76} & \textbf{0.99} & 0.93 & 0.76 & 0.99 \\
        B-cos~\cite{mboehle} & 0.96 & 0.87 & 0.93 & 0.88 & 0.94 & 0.86 & 0.69 & 0.89 & 0.89 & 0.69 & 0.89 \\
        \ourmod\ (ours) & 0.99 & 0.99 & \textbf{0.96} & \textbf{0.98} & \textbf{0.96} & \textbf{0.97} & 0.72 & 0.95 & \textbf{0.97} & 0.72 & 0.95 \\
        \hline
    \end{tabular}
    \caption{Results on the FunnyBirds Framework with a ResNet50 backbone. Note that \textit{Completeness} (Com.) is the mean across Controlled Synthetic Data Check (CSDC), Preservation Check (PC), Deletion Check (DC), and Distractability (D); \textit{Correctness} (Cor.) equals Single Deletion (SD), and \textit{Contrastivity} (Con.) equals Target Sensitivity (TS).}
    \label{tab:fbirds_numbers}
\end{table*}

\begin{figure*}[ht]
    
    \begin{subfigure}{0.49\linewidth}
        \centering
        \includegraphics[width=\linewidth]{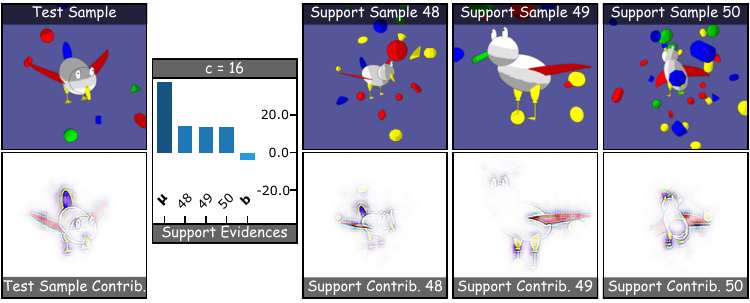}
        \caption{Prediction of the representative FunnyBirds~\citep{hesse2023funnybirds} test sample. The primary location of relevance is the tail of the bird, while the wings, beak, eyes, and legs show a lower contribution. We observe a variety of poses of representative support samples, e.g. a strong focus on the tail and wings in the first support sample, a focus on the legs, eyes, and wings in the second support sample, and a focus on all visible concepts in the last support sample.}
    \label{fig:asdfasdf}
    \end{subfigure}
    \hfill
    \begin{subfigure}{0.49\linewidth}
        \centering
        \includegraphics[width=\linewidth]{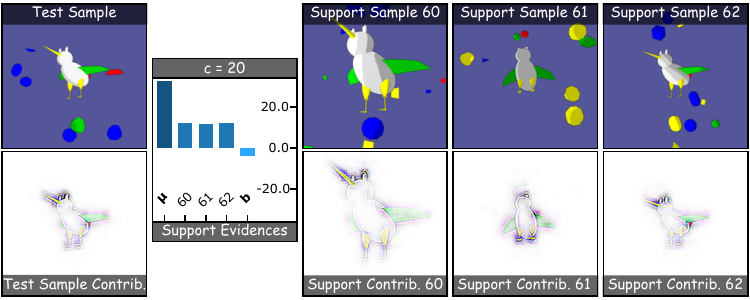}
        \caption{Prediction of another test sample. Similar to (a), all concepts are relevant for the decision making. While the tail appears to be the most important concept for the classification in (a), the eyes seem more important in the specific example. Additionally, the long beak contributes over its whole length, suggesting that the model is sensitive to the appearance of the concept rather than encoding color only.}
    \label{fig:asdfasdffd}
    \end{subfigure}
    \caption{Comparison of Explanations.}
    \label{fig:fbirds_prediction}
\end{figure*}

\subsubsection{CUB-200-2011 - ResNet50}

We trained \ourmod\ on the official splits of CUB-200-2011~\cite{welinder2010caltech} (cropped ROI), achieving an accuracy of 79.8\% (cmp. ProtoPNet 76.1-80.2\% in Tab. 1 (top)~\cite{chen2019looks}).

\clearpage

\subsection{Additional Figures}

\subsubsection{VOC - ResNet50}

\begin{figure}[hb]
    \centering
    \begin{subfigure}{\linewidth}
        \centering
        \includegraphics[width=\linewidth]{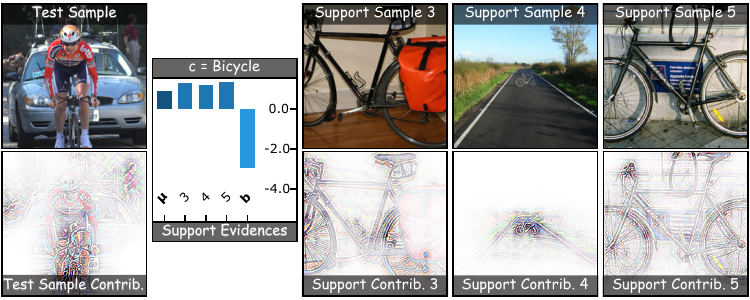}
        \caption{Prediction for the bicycle class.}
    \label{fig:voc_resnet_a}
    \end{subfigure}
    
    \begin{subfigure}{\linewidth}
        \centering
        \includegraphics[width=\linewidth]{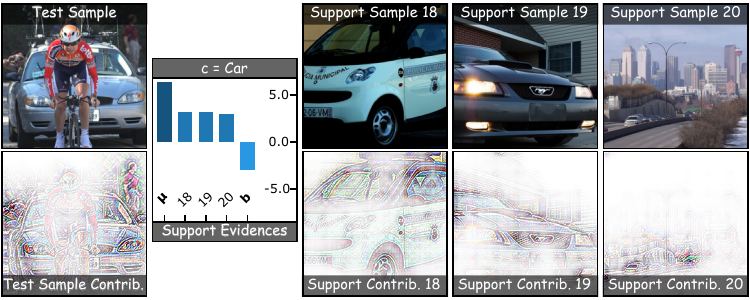}
        \caption{Prediction of the car class. Note the shift in contributions compared to (a).}
    \label{fig:voc_resnet_b}
    \end{subfigure}
    
    \begin{subfigure}{\linewidth}
        \centering
        \includegraphics[width=\linewidth]{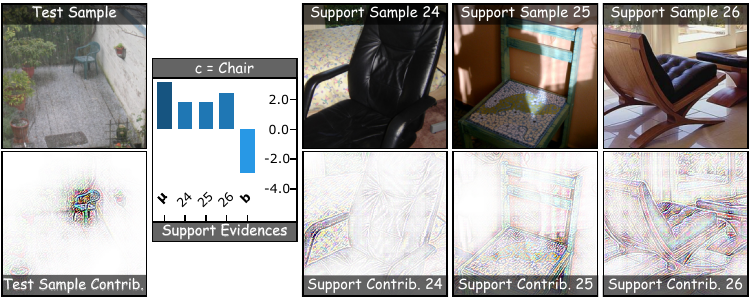}
        \caption{The model correctly found the chair.}
    \label{fig:voc_resnet_c}
    \end{subfigure}
    
    \begin{subfigure}{\linewidth}
        \centering
        \includegraphics[width=\linewidth]{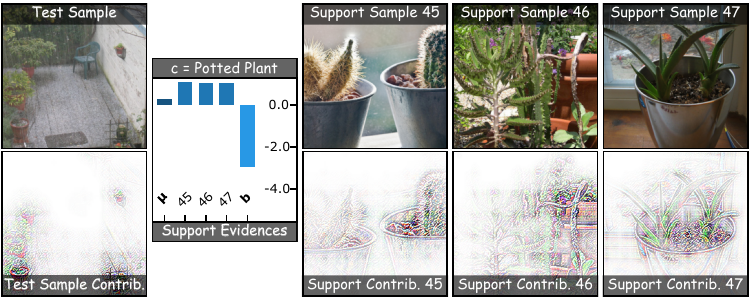}
        \caption{And also the potted plants. Note the shift in contributions compared to (c).}
    \label{fig:voc_resnet_d}
    \end{subfigure}
    \caption{In this and the following figures, they layout introduces marginal changes compared to the main figure. Refer to (a) for the respective explanations. This model was trained with a ResNet50 backbone on VOC.}
    \label{fig:voc_resnet}
\end{figure}

\begin{figure}[hb]
    \centering
    \begin{subfigure}{\linewidth}
        \centering
        \includegraphics[width=\linewidth]{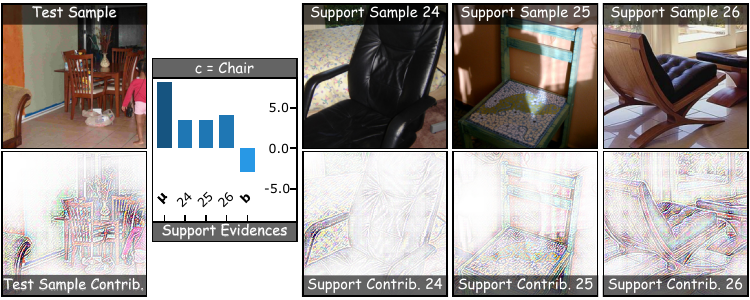}
        \caption{Prediction for the chair class.}
    \label{fig:voc_resnet_wrong_a}
    \end{subfigure}
    
    \begin{subfigure}{\linewidth}
        \centering
        \includegraphics[width=\linewidth]{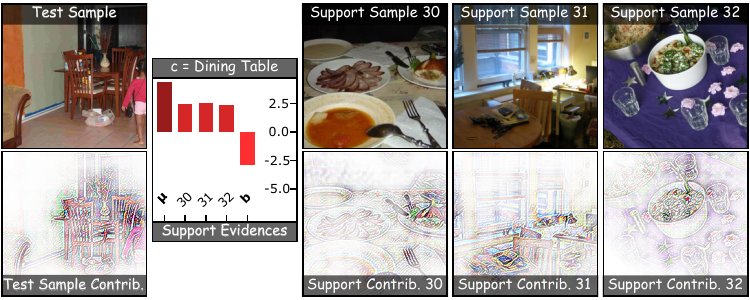}
        \caption{Wrong prediction according to the label of the dining table.}
    \label{fig:voc_resnet_wrong_b}
    \end{subfigure}
    
    \begin{subfigure}{\linewidth}
        \centering
        \includegraphics[width=\linewidth]{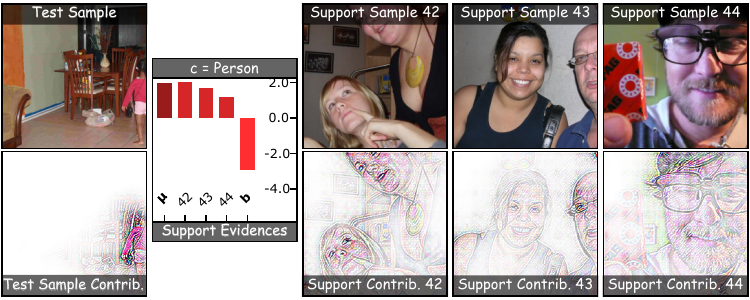}
        \caption{Wrong prediction according to the label of a person.}
    \label{fig:voc_resnet_wrong_c}
    \end{subfigure}
    
    \begin{subfigure}{\linewidth}
        \centering
        \includegraphics[width=\linewidth]{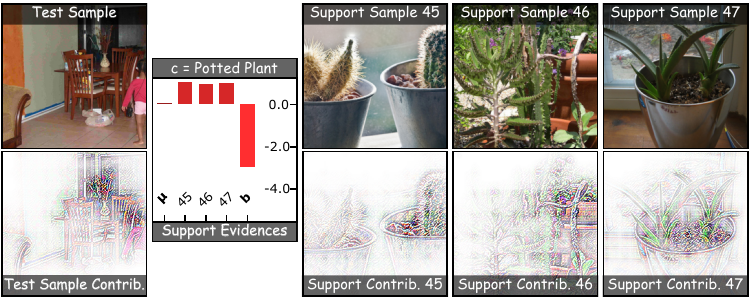}
        \caption{And a wrong prediction, as the flower on the table was mistaken with a potted plant.}
    \label{fig:voc_resnet_wrong_d}
    \end{subfigure}
    \caption{VOC ResNet50}
    \label{fig:voc_resnet_wrong}
\end{figure}

\clearpage
\subsubsection{VOC - VitC}

\begin{figure}[hb]
    \centering
    \begin{subfigure}{\linewidth}
        \centering
        \includegraphics[width=\linewidth]{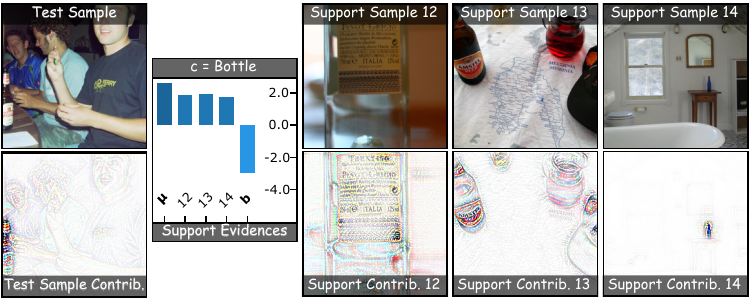}
        \caption{Prediction for the bottle class.}
        \label{fig:voc_vitc_a}
    \end{subfigure}
    
    \begin{subfigure}{\linewidth}
        \centering
        \includegraphics[width=\linewidth]{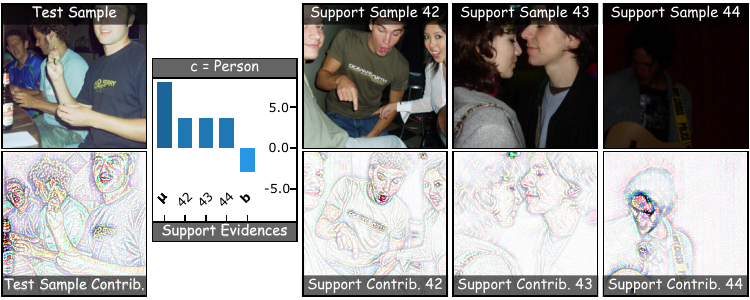}
        \caption{Prediction of the person class.}
        \label{fig:voc_vitc_b}
    \end{subfigure}
    
    \begin{subfigure}{\linewidth}
        \centering
        \includegraphics[width=\linewidth]{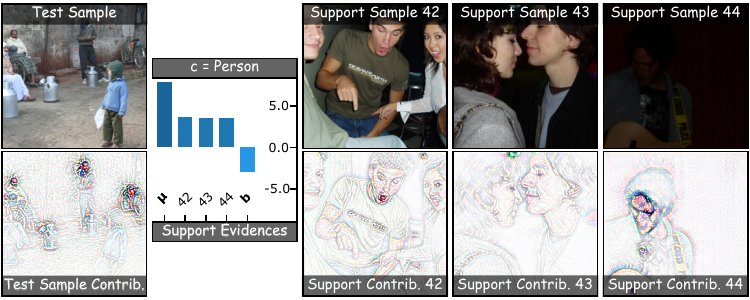}
        \caption{For this test sample, this backbone was the only one not to mistake the milk jugs with bottles.}
        \label{fig:voc_vitc_c}
    \end{subfigure}
    
    \begin{subfigure}{\linewidth}
        \centering
        \includegraphics[width=\linewidth]{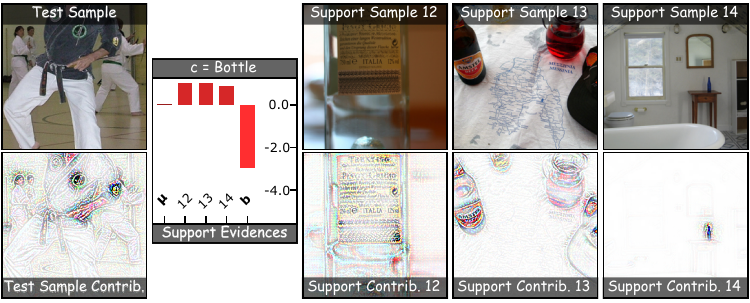}
        \caption{Wrong prediction: The martial arts logo was mistaken with the labels of the first two support samples.}
        \label{fig:voc_vitc_d}
    \end{subfigure}
    
    \begin{subfigure}{\linewidth}
        \centering
        \includegraphics[width=\linewidth]{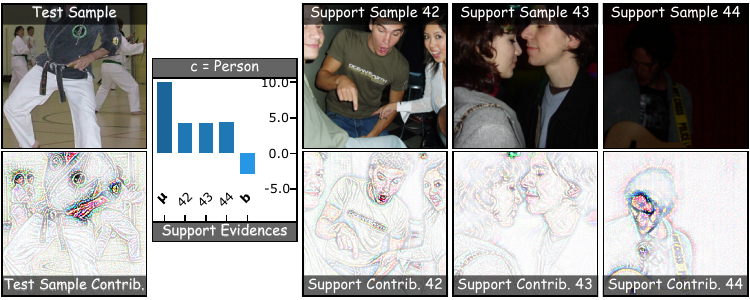}
        \caption{Person was correctly classified.}
        \label{fig:voc_vitc_e}
    \end{subfigure}
    \caption{VOC VitC}
    \label{fig:voc_vitc}
\end{figure}

\clearpage
\subsubsection{Dogs - ResNet50}

\begin{figure}[hb]
    \centering
    \begin{subfigure}{\linewidth}
        \centering
        \includegraphics[width=\linewidth]{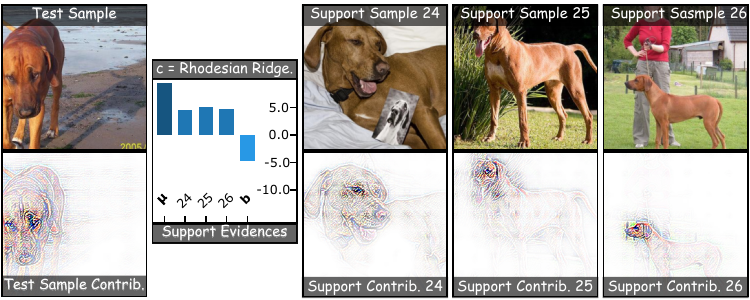}
        \caption{Correct classification. Note the insensitivity to the background.}
    \label{fig:dogs_resnet_0_a}
    \end{subfigure}
    
    \begin{subfigure}{\linewidth}
        \centering
        \includegraphics[width=\linewidth]{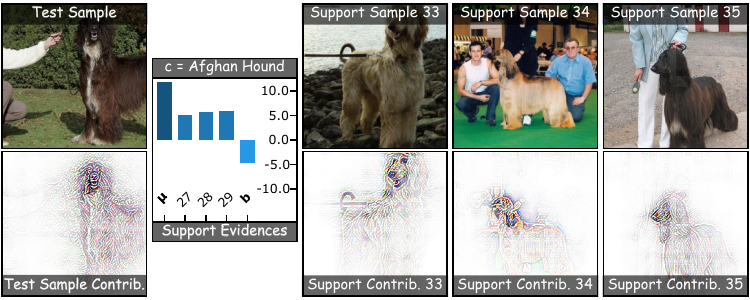}
        \caption{Correct classification. Note the insensitivity to the background.}
    \label{fig:dogs_resnet_0_b}
    \end{subfigure}
    
    \begin{subfigure}{\linewidth}
        \centering
        \includegraphics[width=\linewidth]{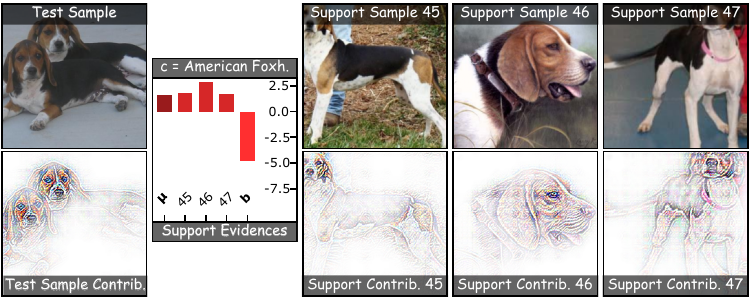}
        \caption{Incorrect classification. The model was very unsure about the class...}
    \label{fig:dogs_resnet_0_c}
    \end{subfigure}
    
    \begin{subfigure}{\linewidth}
        \centering
        \includegraphics[width=\linewidth]{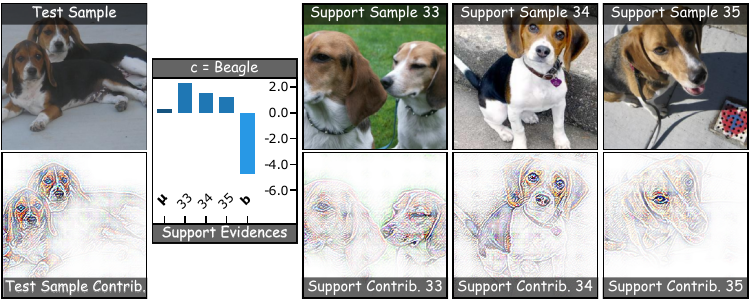}
        \caption{The model found very little evidence for two of the three support samples.}
    \label{fig:dogs_resnet_0_d}
    \end{subfigure}
        \caption{...and the correst class probability suggests that the two breeds are indeed pretty similar.}
    \label{fig:dogs_resnet_0}
\end{figure}

\begin{figure}[hb]
    \centering
    \begin{subfigure}{\linewidth}
        \centering
        \includegraphics[width=\linewidth]{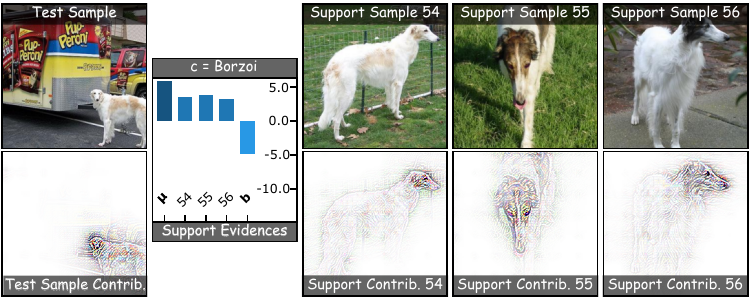}
        \caption{Correct classification. Note the insensitivity to the background.}
    \label{fig:dogs_resnet_1_a}
    \end{subfigure}
    
    \begin{subfigure}{\linewidth}
        \centering
        \includegraphics[width=\linewidth]{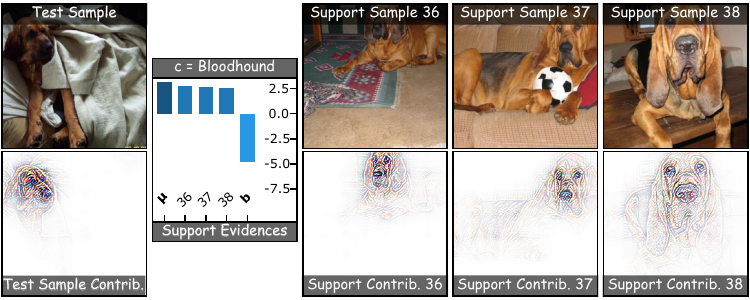}
        \caption{Correct classification. Note the insensitivity to the background.}
    \label{fig:dogs_resnet_1_b}
    \end{subfigure}
    
    \begin{subfigure}{\linewidth}
        \centering
        \includegraphics[width=\linewidth]{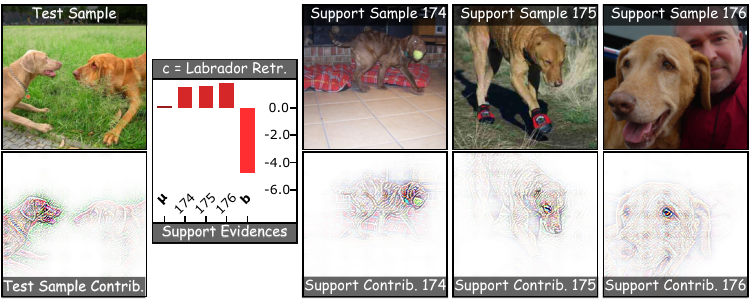}
        \caption{Incorrect classification. The explanations suggest that the dog on the left was prioritized by the network...}
    \label{fig:dogs_resnet_1_c}
    \end{subfigure}
    
    \begin{subfigure}{\linewidth}
        \centering
        \includegraphics[width=\linewidth]{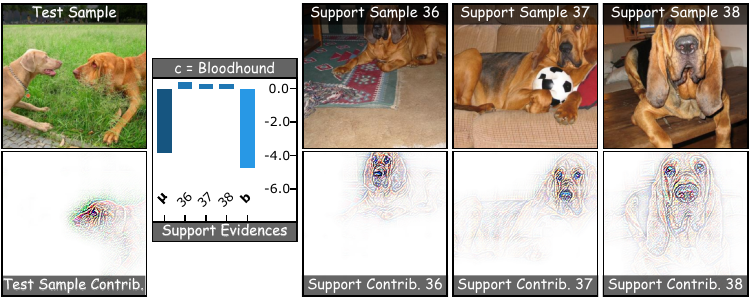}
        \caption{...in contrast to the (arguable) true label. Compared to support samples, the tip of the nose is relatively bright.}
    \label{fig:dogs_resnet_1_d}
    \end{subfigure}
        \caption{Dogs Dataset ResNet50}
    \label{fig:dogs_resnet_1}
\end{figure}

\begin{figure}[hb]
    \centering
    \begin{subfigure}{\linewidth}
        \centering
        \includegraphics[width=\linewidth]{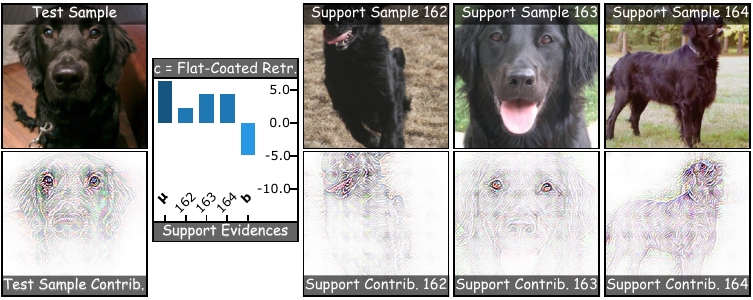}
        \caption{Correct classification. Note how different poses of the support samples lead to different contributions, which is also evident in ...}
    \label{fig:dogs_resnet_2_a}
    \end{subfigure}
    
    \begin{subfigure}{\linewidth}
        \centering
        \includegraphics[width=\linewidth]{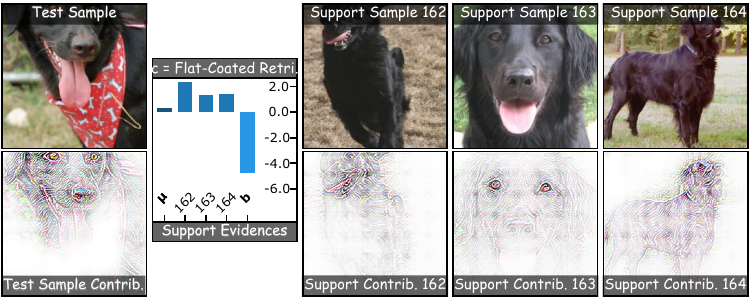}
        \caption{...this figure. Here, the first support sample is most similar to the test image, which is also cropped along the eyeline.}
    \label{fig:dogs_resnet_2_b}
    \end{subfigure}
        \caption{Dogs Dataset ResNet50}
    \label{fig:dogs_resnet_2}
\end{figure}

\clearpage
\subsubsection{RSNA - DenseNet121}

\begin{figure}[hb]

    \begin{subfigure}{\linewidth}
        \centering
        \includegraphics[width=\linewidth]{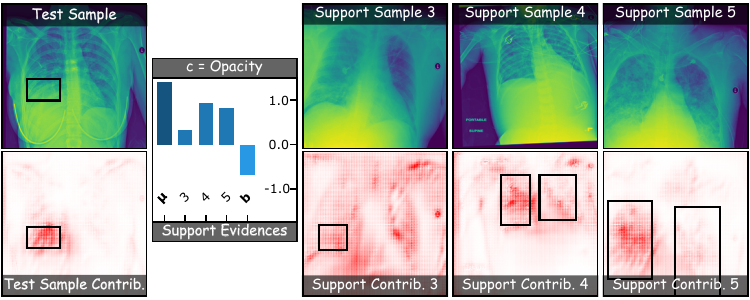}
        \caption{Correct classification of an occlusion present sample.}
    \label{fig:rsna_dense_cor_a}
    \end{subfigure}
    
    \begin{subfigure}{\linewidth}
        \centering
        \includegraphics[width=\linewidth]{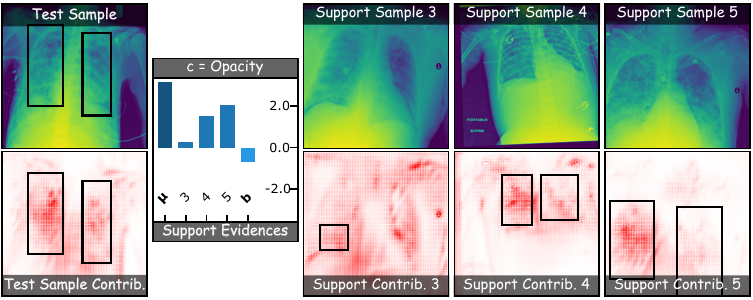}
        \caption{Correct classification of an occlusion present sample.}
    \label{fig:rsna_dense_cor_b}
    \end{subfigure}
    
    \begin{subfigure}{\linewidth}
        \centering
        \includegraphics[width=\linewidth]{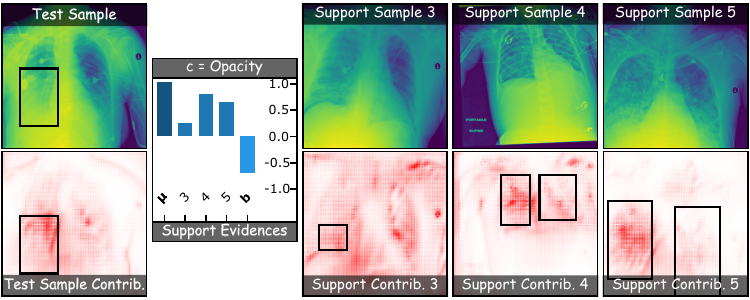}
        \caption{Correct classification of occlusion present sample.}
    \label{fig:rsna_dense_cor_c}
    \end{subfigure}

    \begin{subfigure}{\linewidth}
        \centering
        \includegraphics[width=\linewidth]{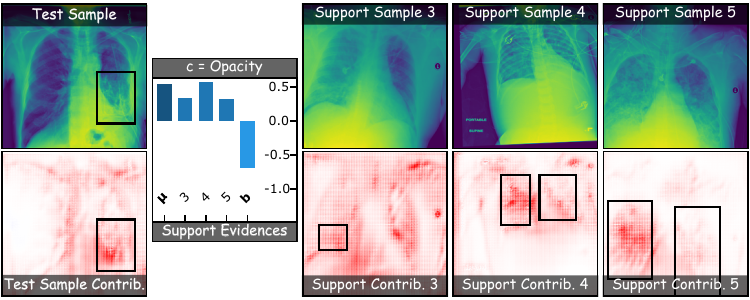}
        \caption{Correct classification of an occlusion present sample.}
    \label{fig:rsna_dense_cor_d}
    \end{subfigure}

    \caption{}
    \label{fig:rsna_dense_cor}
\end{figure}

\begin{figure}[hb]
    \centering
    \begin{subfigure}{\linewidth}
        \centering
        \includegraphics[width=\linewidth]{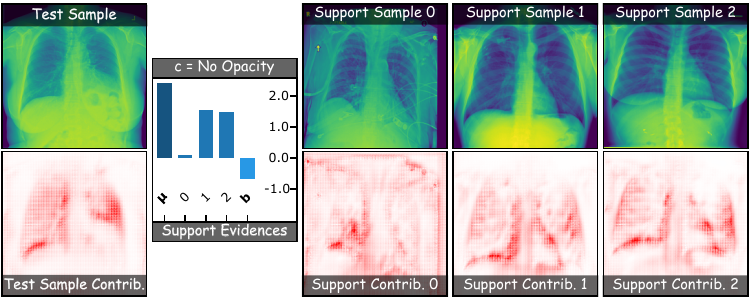}
        \caption{Correct classification of a control sample.}
    \label{fig:rsna_dense_crit_a}
    \end{subfigure}
    
    \begin{subfigure}{\linewidth}
        \centering
        \includegraphics[width=\linewidth]{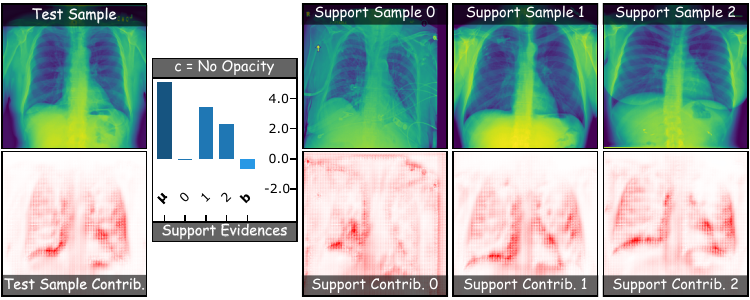}
        \caption{Correct classification of a control sample.}
    \label{fig:rsna_dense_crit_b}
    \end{subfigure}
    \begin{subfigure}{\linewidth}
        \centering
        \includegraphics[width=\linewidth]{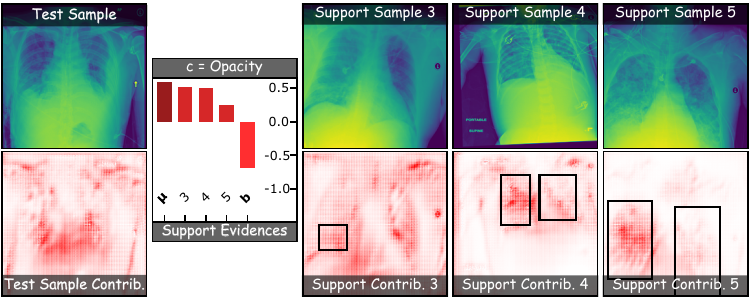}
        \caption{Incorrect classification of a control sample. The model likely mistook the relatively large heart for pathology.}
    \label{fig:rsna_dense_crit_c}
    \end{subfigure}
    
    \begin{subfigure}{\linewidth}
        \centering
        \includegraphics[width=\linewidth]{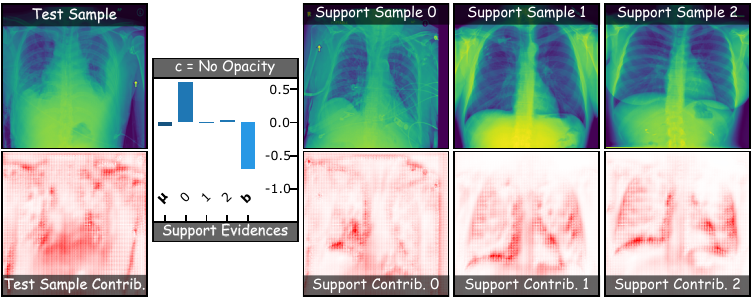}
        \caption{The model found very little evidence for two of the three support samples.}
    \label{fig:rsna_dense_crit_d}
    \end{subfigure}
    \caption{}
    \label{fig:rsna_dense_crit}
\end{figure}

\begin{figure*}[ht]
    \centering
    \includegraphics[width=0.8\textwidth]{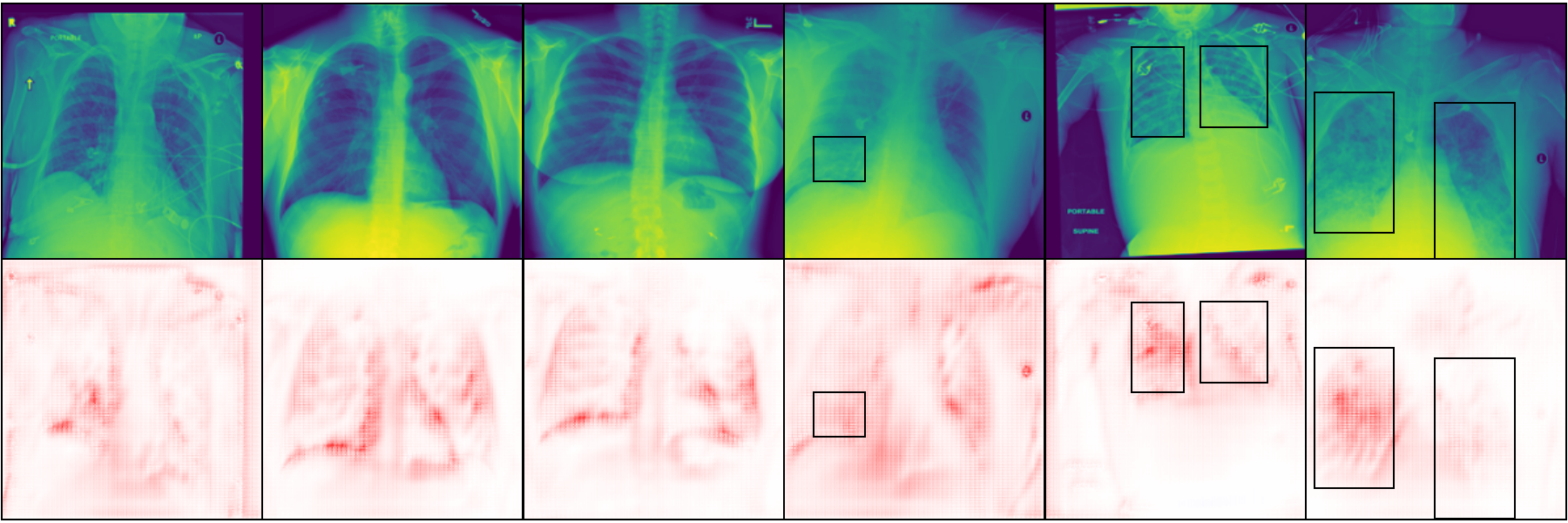}
    \caption{Prototypes of \ourmod\ trained with a DenseNet121 backbone on RSNA. Bounding boxes indicate where medical staff found occlusion of the lungs. The first three columns are the support samples of controls, the next three are support samples with occlusion present.}
    \label{fig:rsna_densenet_protos}
\end{figure*}

\begin{figure*}[!hb]
    \centering
    \includegraphics[width=0.8\textwidth]{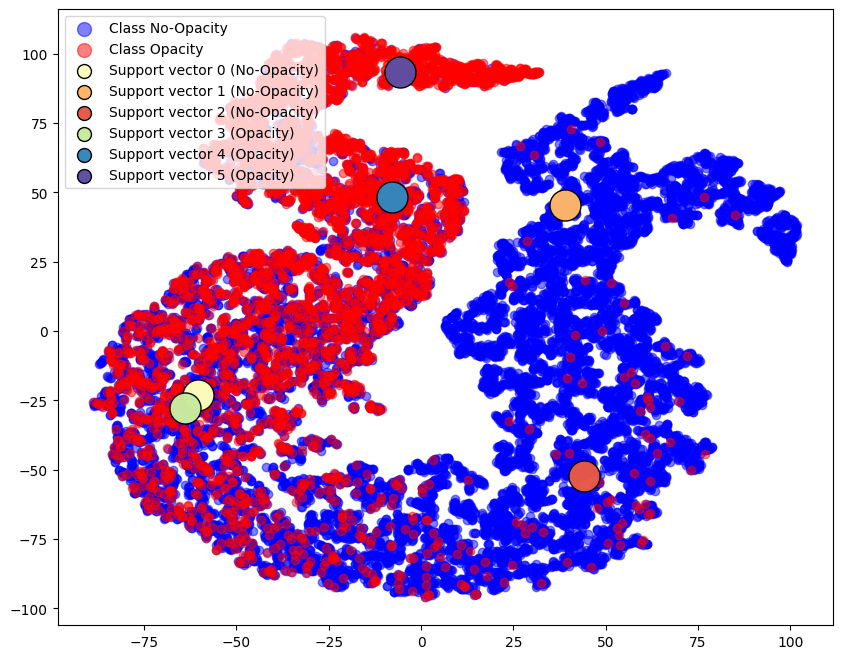}
    \caption{t-SNE plot of the latent vectors of the RSNA training set. Red denotes samples of the class "opacity" and blue the class "no opacity". Support vector 0 and support vector 3 are very close in the projected latent visualization.}
    \label{fig:tsne_rsna}
\end{figure*}

\subsubsection{VOC - DenseNet121: Prototypes and B-cos Similarities}

\begin{figure*}[!hb]
    \centering
    \includegraphics[width=\textwidth]{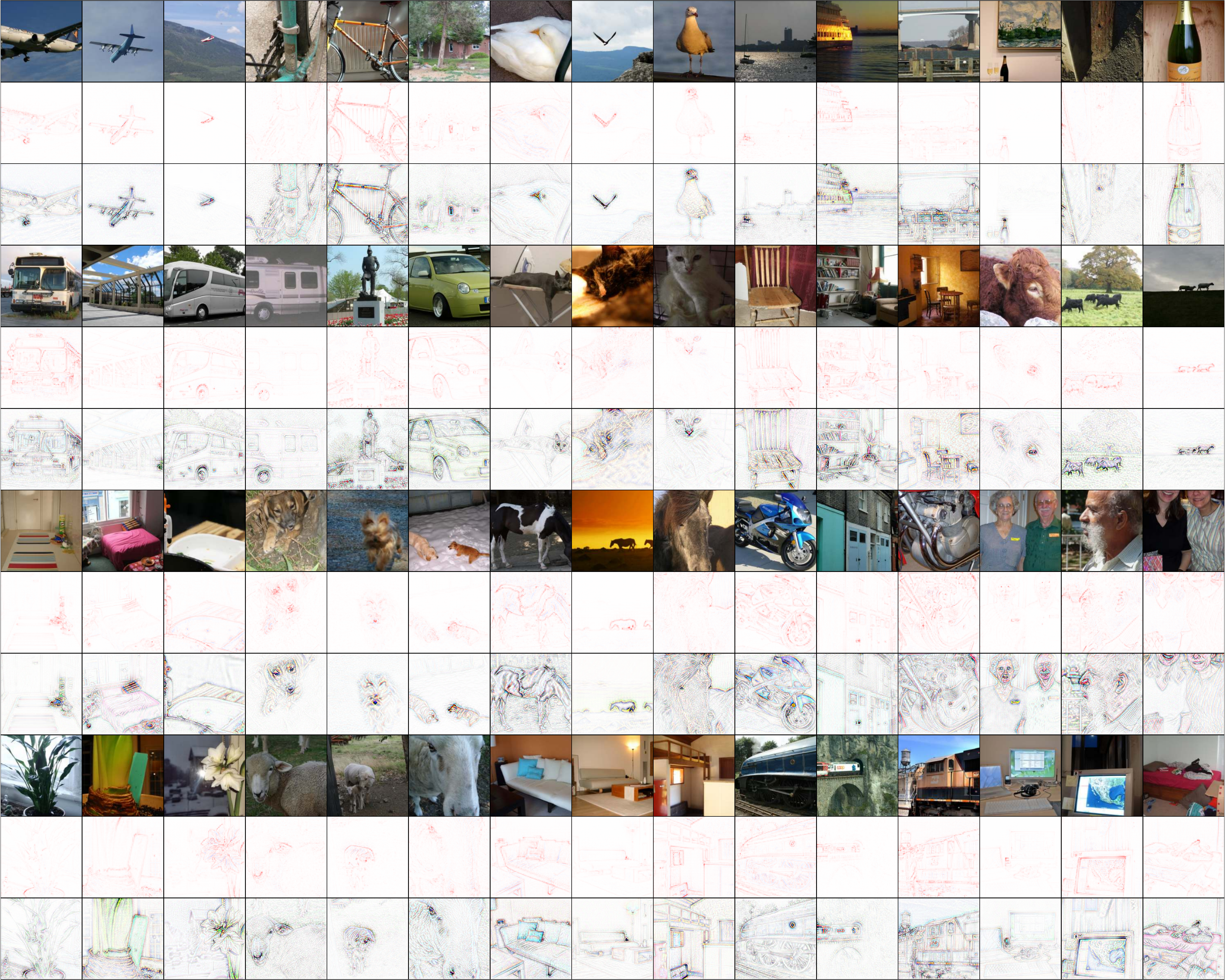}
    \caption{Global explanations for \ourmod\ trained on Pascal VOC with a DenseNet121 backbone. The respective second rows show the raw contribution maps $\phi^L_{j=c}$, in which red denotes positive contributions and blue (absent) negative contributions. The respective third columns present RGBA explanations.}
    \label{fig:voc_protos}
\end{figure*}

\begin{figure*}[!hb]
    \centering
    \includegraphics[width=0.8\textwidth]{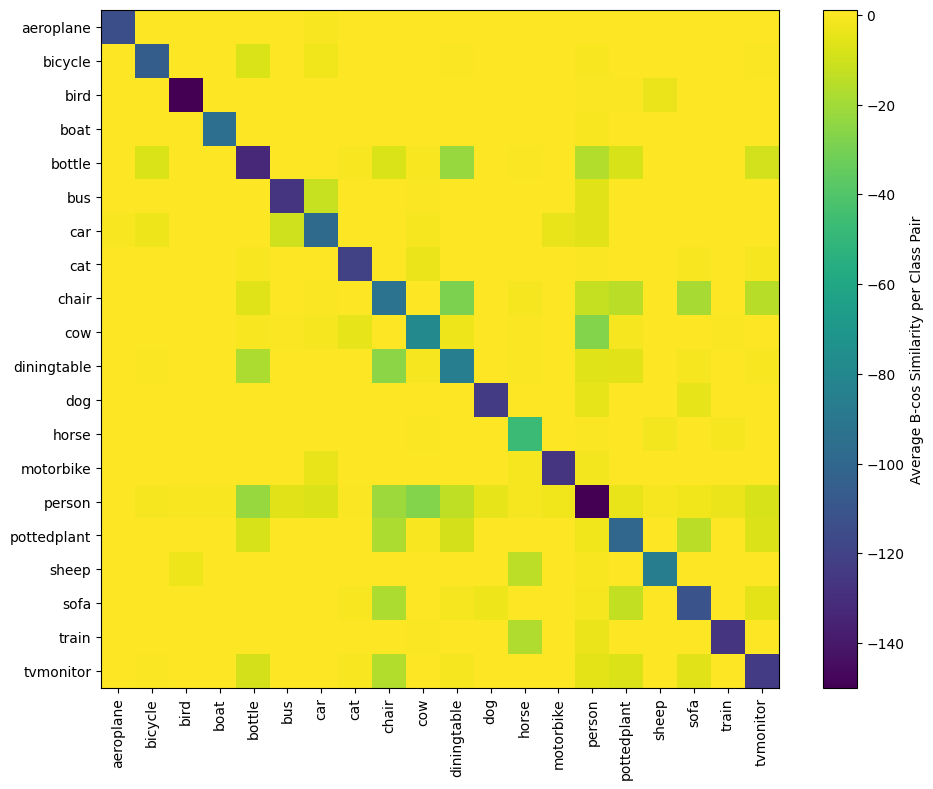}
    \caption{Heatmap of Intra-class and Inter-class B-cos Similarities for \ourmod\ trained on Pascal VOC with a DenseNet121 backbone.}
    \label{fig:voc_heatmap}
\end{figure*}

\end{document}